\documentclass{amsart}
\usepackage[utf8]{inputenc}
\usepackage{graphicx}
\usepackage[margin=1in]{geometry}
\usepackage{multirow}
\usepackage{longtable}
\usepackage{lscape}
\usepackage{tablefootnote}
\usepackage{subcaption}
\usepackage{dsfont}
\usepackage{mathrsfs}
\usepackage{amssymb}
\usepackage{enumerate}
\usepackage{graphicx}
\usepackage{float}
\usepackage{bbm}
\usepackage{amsmath}
\usepackage{comment}
\usepackage{hyperref}
\usepackage{listings}
\usepackage{color}
\usepackage{ulem}
\usepackage{algorithm,algorithmicx,algpseudocode}
\graphicspath{{figures/}}
\usepackage{xcolor}

\makeatletter
\newcommand{\addresseshere}{%
  \enddoc@text\let\enddoc@text\relax
}
\makeatother

\newcommand{\curly}[1]{\left\{ #1 \right\}}

\hypersetup{
	colorlinks,
	citecolor=blue,
	filecolor=blue,
	linkcolor=blue,
	urlcolor=blue,
	hyperfootnotes=false
}

\newtheorem{theorem}{Theorem}[section]
\newtheorem{proposition}[theorem]{Proposition}

\newtheorem{corollary}[theorem]{Corollary}
\theoremstyle{definition}

\newtheorem{remark}{Remark}
\newtheorem{definition}{Definition}

\newcommand{\R}{\mathbb{R}}

\newcommand{\N}{\mathbb{N}}

\newcommand{\rank}{{\rm rank\,}}

\newcommand{\argmin}{\text{argmin}}

\newcommand{\diag}{\textnormal{diag}}

\title{On Matrix Factorizations in Subspace Clustering}

\author{Reeshad Arian}
\address{Department of Mathematics, University of Arizona, Tucson, AZ}
\email{rarian@email.arizona.edu} 

\author{Keaton Hamm}
\address{Department of Mathematics, University of Texas at Arlington, Arlington, TX}
\email{keaton.hamm@uta.edu}



\begin{document}

\begin{abstract}

This article explores subspace clustering algorithms using CUR decompositions, and examines the effect of various hyperparameters in these algorithms on clustering performance on two real-world benchmark datasets, the Hopkins155 motion segmentation dataset and the Yale face dataset.  Extensive experiments are done for a variety of sampling methods and oversampling parameters for these datasets, and some guidelines for parameter choices are given for practical applications.
    
\end{abstract}

\maketitle


\section{Introduction}

While most modern data is high-dimensional and large-scale, it is increasingly understood that in actuality, typical data exhibits significant structure rather than inhabiting the whole ambient space.  In particular, many data sets are well-modelled to comprise low-dimensional structures such as linear subspaces or non-linear manifolds. The manifold hypothesis (that the data lives in a neighborhood of a low-dimensional embedded manifold) \cite{fefferman2016testing} is oft-utilized to good effect in Machine Learning applications, and is considered as a potential explanation for adverse behavior of neural network classifiers which are susceptible to small adversarial attacks which drastically change the classification \cite{szegedy2014intriguing}; said another way, such classifiers are not input stable.

The \textit{union of subspaces model} for data is that data vectors lie in $\mathscr{U}:=\bigcup_{i=1}^L S_i\subset\R^n$, where each $S_i$ is a linear (or affine) subspace of $\R^n$.  While seemingly specific, the union of subspaces model describes data from a substantial number of applications, including motion segmentation, facial recognition, and cryo-electron microscopy.  We will discuss specifics of applications later, but for now we also note that $\Sigma_s^n$, the set of $s$--sparse signals in $\R^n$, is the union of all $s$--dimensional coordinate subspaces, and that sparse signal reconstruction has been an active and influential field since the advent of compressed sensing \cite{donoho2006compressed,foucart2013invitation}.

The \textit{subspace clustering problem} is: given data $\{x_i\}_{i=1}^m\subset\R^n$ coming from a union of subspaces model as above, find a mapping $\Phi:\{x_i\}_{i=1}^m\to\{1,\dots,L\}$ such that $\Phi(x_i)=j$ if $x_i\in S_j$, and is $0$ otherwise.  There are various solutions to this problem \cite{vidal2011subspace}, but we highlight here those based on matrix factorizations \cite{aldroubi2019cur,aldroubi2018similarity,costeira1998multibody,rsimpaper} and on the self-expressive property of subspace data \cite{aldroubi2012nearness,sparsesubspaceclustering,elhamifar2013sparse,liu2010robust,liu2012robust}.  More details follow in Section \ref{SEC:Theory}.

In \cite{aldroubi2019cur,aldroubi2018similarity}, Aldroubi et al.~proposed the use of CUR decompositions for robust algorithms to solve the subspace clustering problem.  In \cite{aldroubi2019cur}, the proposed algorithm contained many hyperparameters which could be changed, but were untested therein. The purpose of this work is to consider variations of the robust CUR similarity matrix algorithm of \cite{aldroubi2019cur}, examine the effect of various parameters on clustering outcome, and extend the analysis to other datasets than the lone one considered there.

\section{Preliminaries}

We consider real-valued data matrices $X\in\R^{m\times n}$, and let $[n]:=\{1,\dots,n\}$ for any $n\in\N$.  Any matrix has a Singular Value Decomposition $X=U\Sigma V^T$ in which $U\in\R^{m\times m}$ and $V\in\R^{n\times n}$ are orthogonal, and $\Sigma\in\R^{m\times n}$ has nonzero entries only along its main diagonal.  Columns of $U$ (resp. $V$) are left (resp. right) singular vectors of $X$, and the diagonal entries of $\Sigma$ are its singular values, which we denote $\{\sigma_i\}_{i=1}^{\min\{m,n\}}$, or $\sigma_i(X)$ if need be.  Singular values are assumed to be ordered so that $\sigma_1\geq\sigma_2\geq\dots$, and note that if $\rank(X)=k$, then $\sigma_j=0$ for all $j>k$.  Given a rank $k$ matrix $X$, we may write $X=U_k\Sigma_kV_k^T$, where $U_k$ and $V_k$ contain only the first $k$ columns of $U$ and $V$, respectively, and $\Sigma_k$ is the $k\times k$ submatrix of $\Sigma$ containing only the positive singular values along its diagonal. This decomposition is called the compact SVD of $X$. If $X$ is not rank $k$, then we call $X_k=U_k\Sigma_kV_k^T$ the truncated SVD of $X$ of order $k$, which is the best rank $k$ approximation of $X$ (in any Schatten $p$--norm) by the Eckart--Young--Mirsky Theorem. For any $X$ with rank $k$, its Moore--Penrose pseudoinverse, denoted by $X^\dagger$, is given by $X^\dagger = V_k\Sigma_k^{-1}U_k^T$. Finally, $\textnormal{abs}(X)$ will denote the matrix whose $i,j$--th entry is $|X_{ij}|$.

The general case of the subspace clustering problem in which the model $\mathscr{U}$ consists of arbitrarily many subspaces of varying dimensions which are allowed to overlap nontrivially is quite challenging. Simpler cases are when the subspaces are \textit{disjoint} ($S_i\cap S_j=\{0\}$, $i\neq j$) or \textit{independent} ($\dim(\sum_{i=1}^LS_i) = \sum_{i=1}^L\dim(S_i)$, where $\sum S_i$ denotes Minkowski sum). The latter case has the most structure to exploit and has been more thoroughly explored; indeed, there are many guarantees of correct clustering for independent subspaces.

We will not dwell much on how data is sampled or obtained from each subspace. However, we will make a typical assumption that data in each subspace is \textit{in general position} (also called \textit{generic}).  Given a subspace $S\subset\R^m$ of dimension $d$, points $Y=\{y_i\}_{i=1}^t\subset S$ are in general position if any collection of $d$ points from $Y$ form a basis for $S$. This assumption is not restrictive; if points are sampled uniformly from a ball in the subspace, then they are almost surely in general position.

\section{Theoretical Background}\label{SEC:Theory}

The \textit{self-expressive property} of data is that one can write $X=XZ$ for some matrix $Z$ with $\diag(Z)=0$ \cite{sparsesubspaceclustering}.  In particular, this assumes that any data vector $x_i$ in $X$ can be written as a linear combination of other data vectors in $X$ \textit{excluding} $x_i$ itself. In practice, $Z$ is typically chosen to minimize $\|X-XZ\|_F$ or some similar penalized objective function. 

Several matrix factorization approaches to subspace clustering have been proposed in which one factorizes $X=AZ$ for some $A$ (for example, coming from the compact SVD, QR decomposition, CUR decomposition, or something else).  Both matrix factorization techniques and those that arise from the self-expressive property treat $Z$ or some suitable matrix coming from it as a \textit{similarity matrix} for the subspace data $X$.  That is, $Z_{ij}$ should be relatively large if $x_i$ and $x_j$ are in the same subspace and relatively small (ideally $0$) if they are in different subspaces.  These ideas are closely tied to similarity matrix techniques for general clustering tasks, e.g., spectral clustering \cite{von2007tutorial}.

Inherent in these methods is that the similarity matrix $\Xi$ is treated as the adjacency (weight) matrix of a graph whose vertices are the data points $x_i$, and success of the clustering method is dependent upon the graph structure and properties of the graph Laplacian.  Roughly speaking, one desires that $\Xi$ consist of $L$ (approximately) disjoint connected components which correspond to the $L$ subspaces in the union of subspaces model.  We refer the reader to the discussion of von Luxburg \cite{von2007tutorial} which elucidates the strong connection between spectral clustering and several well-known graph cut problems.

\subsection{Clustering Matrices}

More than similarity matrices, which compute some measure of simlarity between data points, we can ask for an idealized matrix which tells us the subspace clusters directly. In particular, we make the following definition.

\begin{definition}
Let $X\in\R^{m\times n}$ have columns drawn from a union of subspaces $\mathscr{U}=\bigcup_{i=1}^L\subset\R^n$.  A matrix $\Xi\in\R^{n\times n}$ is a \textit{clustering matrix} for $X$ provided $\Xi_{ij}\neq0$ if and only if $x_i$ and $x_j$ are in the same subspace.
\end{definition}

Evidently, finding an actual clustering matrix from union of subspace data is challenging, especially in the presence of noise, but nonetheless there are various ways of doing so as the following theoretical results will show. 

To begin, we describe two types of methods for finding clustering matrices: sparse optimization methods and direct matrix factorization methods.  To begin, we present two optimization methods -- Sparse Subspace Clustering (SSC) \cite{sparsesubspaceclustering} and Low-rank Representation (LRR) \cite{liu2012robust} -- the first of which seeks a representation $X=XZ$ with a sparse coefficient matrix $Z$, while the second seeks a spectrally sparse representation of the same form (i.e., $\rank(Z)$ is small).

The Sparse Subspace Clustering problem is 
\begin{equation}\label{EQN:SSC}
\min_{Z\in\R^{n\times n}} \|Z\|_1 \quad \textnormal{subject to} \quad X=XZ.
\end{equation}
The Low-Rank Representation problem is
\begin{equation}\label{EQN:LRR}
\min_{Z\in\R^{n\times n}} \|Z\|_\ast \quad \textnormal{subject to} \quad X=XZ.
\end{equation}
Here $\|Z\|_*=\sum_i \sigma_i(Z)$ is the nuclear norm, or equivalently the $\ell_1$ norm of the singular values of $Z$. 

The problems \eqref{EQN:SSC} and \eqref{EQN:LRR} are actually convex relaxations of the combinatorial optimization problems that more naturally describe both SSC and LRR, in which one desires to minimize $\|Z\|_0$ (the number of nonzero entries of $Z$) and $\rank(Z)$, respectively. The $\ell_1$--norm and nuclear norm are the convex envelopes of the $\ell_0$--``norm" and rank (which is the $\ell_0$--``norm" of the singular values of $Z$), respectively.

In the presence of noise, both of these optimization problems have natural extensions, but we refer the interested reader to the references above for more information. However, for the reader's convenience we mention the following which shows that both of these formulations yield a clustering matrix in the ideal case.

\begin{theorem}[\cite{elhamifar2013sparse,liu2012robust}]\label{THM:SSCLRR}
Let $X\in\R^{m\times n}$ have columns drawn from a union of independent subspaces $\mathscr{U}$ such that the data from each subspace is in general position. Let $Z$ be the solution to either Sparse Subspace Clustering or Low-rank Representation optimization problems \eqref{EQN:SSC} or \eqref{EQN:LRR}, respectively.  Then $Z$ is a clustering matrix for $X$.
\end{theorem}

We now turn to methods which arise from directly trying to factorize the matrix $X$.  We begin by a fairly general result of Aldroubi et al. \cite{aldroubi2018similarity}.

\begin{theorem}[{\cite[Theorem 2]{aldroubi2018similarity}}]\label{THM:ACKS}
Let $X\in\R^{m\times n}$ have columns drawn from a union of independent subspaces $\mathscr{U}$ such that the data from each subspace is in general position.  Let $B\in\R^{m\times r}$ have columns which also lie in $\mathscr{U}$ and such that the columns of $B$ form a basis for the column space of $X$.  Then $X=BP$ for some unique $P\in\R^{r\times n}$, and $\Xi:=\textnormal{abs}(P^TP)^{d_{\max}}$ is a clustering matrix for $X$.
\end{theorem}

This theorem is not practically useful given that one must exactly choose a basis from the union of subspace data to yield the clustering matrix.  However, it provides a useful theoretical framework from which one can build. In particular, it inspired a more general result below. To state this, we must first discuss CUR decompositions.

\begin{proposition}\label{PROP:CURDecomp}
Let $X\in\R^{m\times n}$ have rank $r$. Let $I\subset[m]$, $J\subset[n]$. with $C=X(:,J)$, $U=X(I,J)$, and $R=X(I,:)$.  If $\rank(U)=\rank(X)$, then $X=CU^\dagger R$.  
\end{proposition}

Proposition \ref{PROP:CURDecomp} is essentially folklore. For a history and characterization of CUR decompositions for low-rank matrices, the reader is encouraged to consult \cite{hamm2020perspectives}.

\begin{theorem}[{\cite[Theorem 2]{aldroubi2019cur}}]\label{THM:CURClustering}
Let $X\in\R^{m\times n}$ have columns drawn from a union of independent subspaces $\mathscr{U}$ such that the data from each subspace is in general position.  Let $I\subset[m]$ and $J\subset[n]$ be such that $C=X(:,J)$, $U=X(I,J)$, and $R=X(I,:)$ yield a CUR decomposition of $X$ (i.e., $X=CU^\dagger R$).  Then $\Xi:=\textnormal{abs}((U^\dagger R)^TU^\dagger R)^{d_{\max}}$ is a clustering matrix for $X$.
\end{theorem}

Interestingly, the CUR clustering matrix theorem above provides an overarching framework which encompasses Low-Rank Representation as well as the classical SVD-based clustering matrix method of Costeira and Kanade \cite{costeira1998multibody}.

\begin{corollary}\label{COR:SIM}
Let $X\in\R^{m\times n}$ have columns drawn from a union of independent subspaces $\mathscr{U}$ such that the data from each subspace is in general position.  If $X=U_r\Sigma_rV_r^T$ is the compact SVD of $X$, then $S:=\textnormal{abs}(V_rV_r^T)^{d_{\max}}$ is a clustering matrix for $X$.  
\end{corollary}

\begin{remark}
The use of the compact SVD for subspace clustering was first considered by Costeira and Kanade \cite{costeira1998multibody} for the task of motion segmentation. They dub the matrix $V_rV_r^T$ the \textit{shape interaction matrix}.  We note for the reader that this is not guaranteed to be a clustering matrix for $X$; however it is almost surely a clustering matrix assuming data is drawn from each subspace uniformly at random (see \cite{aldroubi2018similarity} for precise statements to these effects).  The matrix power $d_{\max}$ fixes this gap, but is not typically advised in practice as matrix powers can inflate the effect of noise on the similarity matrix.
\end{remark}

\subsection{Sampling methods for CUR decompositions}

There are various ways of selecting the row and column indices to form a CUR decomposition. We will not discuss them all here, but will focus on a range of random methods and one deterministic method.  

One of the most common random sampling procedures for CUR decompositions is to form probability distributions over the columns and rows and sample from them with or without replacement. There are variants which sample according to Bernoulli trials on each column and row, but the theoretical guarantees (with respect to sampling order and approximation of the truncated SVD by CUR approximations) are typically worse in these cases.  We focus on the three most common distributions used in the literature: uniform, column/row length, and leverage scores.  For column selection, the distributions are given as follows:
\begin{equation}\label{EQ:Probabilities}q_j^{\textnormal{unif}}:=\frac1n,\quad q_j^{\textnormal{col}}:=\frac{\|X(:,j)\|_2^2}{\|X\|_F^2},\quad q_j^{\textnormal{lev,k}}:=\frac1k \|V_k(j,:)\|_2^2.\end{equation}
Row selection distributions are defined analogously, and denoted by $p_i^{\textnormal{unif}}, p_i^{\textnormal{row}},$ and  $p_i^{\textnormal{lev},k}$.  

It should be noted that the computation required to compute the distributions themselves increases sharply from left to right, given that there is no computational cost to computing the uniform distribution, whereas the leverage scores require $O(mnk)$ flops to compute.  On the other hand, the information used about the underlying matrix increases from left to right as well.  It should thus be of no surprise that leverage score sampling typically yields the best results in theory \cite{mahoney2009cur} followed by column length and then uniform.  However, there is a notable exception to the quality of uniform sampling: it is known to provide quality submatrices when $X$ has small incoherence \cite{chiu2013sublinear}.

For deterministic column sampling, we consider the \textit{Discrete Empirical Interpolation Method (DEIM)} which was originally considered by Gu and Eisenstat \cite{gu1996efficient} in relation to rank-revealing QR, and subsequently for CUR decompositions by Sorensen and Embree \cite{sorensen2016deim}.  We refer the readers to the latter reference for the complete algorithm, but essentially DEIM sampling first chooses the column corresponding to the index of the largest (in absolute value) entry of the first right singular vector, forms an oblique projector onto that index, subtracts the projection of the second singular vector from itself, and then repeats this process until the desired number of columns is chosen.

\subsection{Noise Effects}

\begin{figure}[h]
\centering
\includegraphics[width=.32\textwidth]{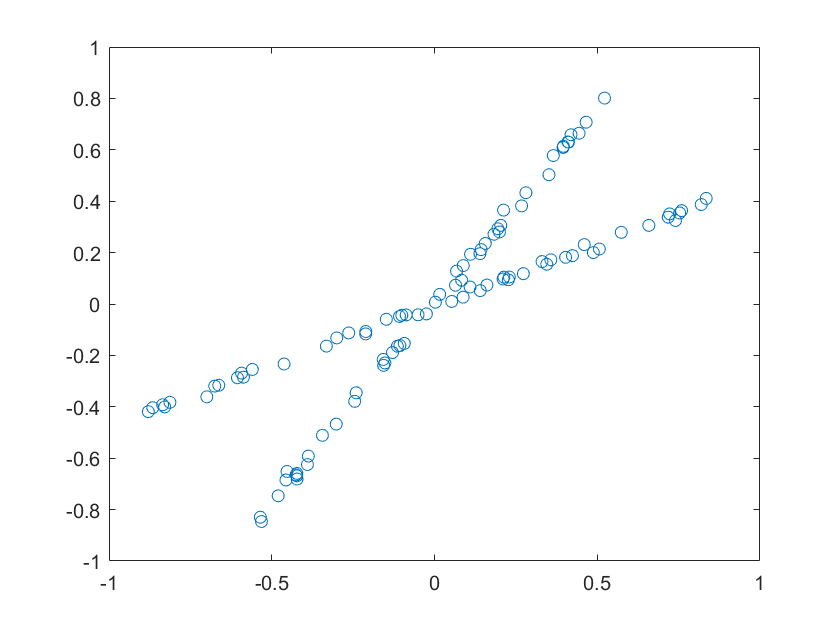} \includegraphics[width=.32\textwidth]{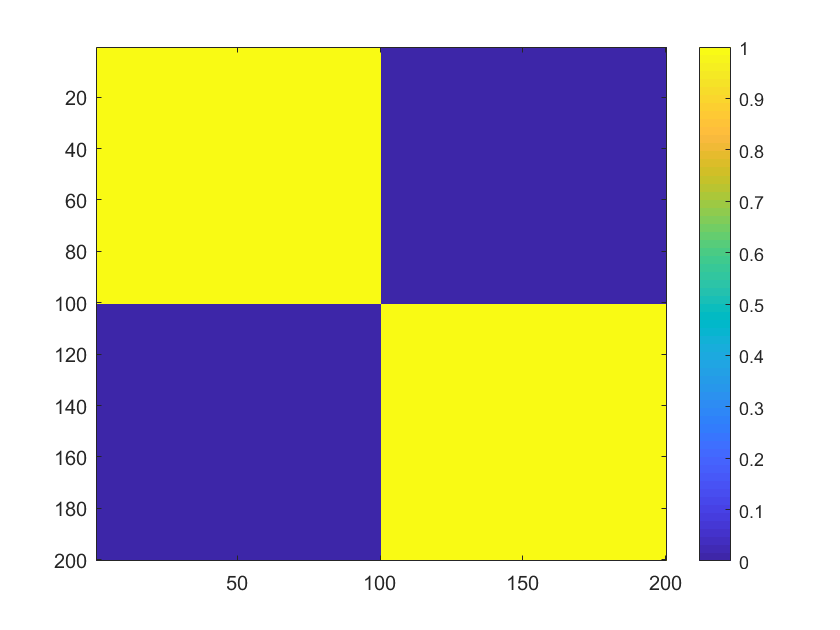}\includegraphics[width=.32\textwidth]{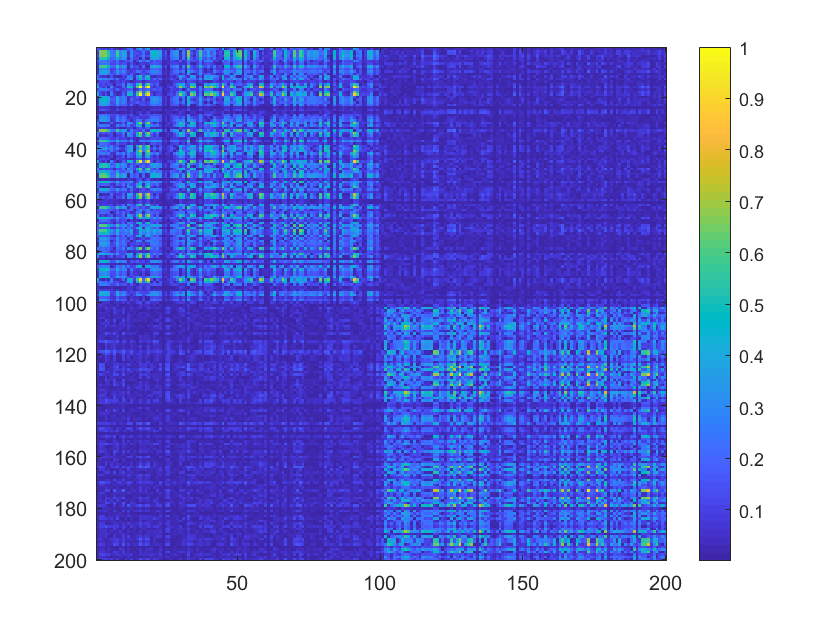}
\caption{Here, the subspace data $X$ is drawn uniformly from two random lines in $\R^2$, and the noise $E$ has i.i.d. $\mathcal{N}(0,10^{-2})$ entries. Data is ordered so that the first 100 points come from the first line, and the next 100 points from the second.  The noisy data set is shown on the left, while a pure clustering matrix is shown in the center and the shape interaction matrix $\textnormal{abs}(V_2V_2^T)$ is shown on the right.}\label{FIG:ToyData}
\end{figure}

In practice, we actually observe $D = X+E$, where columns of $X$ are drawn from a union of subspaces model $\mathscr{U}$ and $E$ is a matrix of noise.  It is important to note that classical methods like the SVD-based method of \cite{costeira1998multibody} are insufficient to overcome additive noise even with small magnitude. To illustrate this, we show a simple toy example in Figure \ref{FIG:ToyData}.

At first glance, the noisy clustering matrix on the right of Figure \ref{FIG:ToyData} appears sufficiently clear; however, recalling that the data is ordered, we notice that there are deep blue valleys near the edges of the top left and bottom right blocks. These indicate that the noise has caused points on opposite ends of the same line to be clustered differently.  Ambiguity near the origin is understandable and perhaps unavoidable, but misclustering points far away on the same subspace is undesirable. The next section will discuss robust versions of the proposed clustering matrices discussed above.

\subsection{Robust Similarity Matrices}

Ji, Salzmann, and Li \cite{rsimpaper} propose a simple modification of the shape interaction matrix which is robust to noise. Inspired by this, Aldroubi et al. \cite{aldroubi2019cur} proposed a robust CUR similarity matrix procedure which achieved the best performance to date on the Hopkins155 motion segmentation data set \cite{tron2007benchmark}. The algorithm proposed therein contained many parameters, and as it is the purpose of this work to explore the effect of some of these, we now describe the algorithm in detail.

CUR decompositions are well-known for their flexibility due to the fact that a low-rank matrix admits many possible CUR decompositions by choosing different index sets $I$ and $J$ to form the constituent matrices.  This flexibility can be used to great advantage in the application of Theorem \ref{THM:CURClustering} when the data $X$ contains noise or outliers. The main idea, explored in \cite{aldroubi2019cur}, is to form $N$ noisy clustering matrices for $X$ from $N$ distinct CUR approximations followed by some sort of aggregation and thresholding procedure, whereupon spectral clustering is applied to this denoised clustering matrix to return the final cluster labels. We reproduce here a slightly generalized variant of this algorithm, which we call Robust CUR Similarity Matrix algorithm (or RCUR for short), below.

\begin{algorithm}
\caption{Robust CUR Similarity Matrix (RCUR)}\label{ALG:RCUR}
\begin{algorithmic}[1]
\Statex \textbf{Input:} $D\in\R^{m\times n}$ with $D=X+E$ where columns of $X$ are from a union of linear subspaces $\mathscr{U}=\bigcup_{i=1}^LS_i$ and $E$ is noise, minimum and maximum rank parameters ($r_{\min}$, $r_{\max}$), probability distributions for each rank parameter ($\{p^{(r)}_i\}_{i=1}^m$, $\{q^{(r)}_j\}_{j=1}^n$), and size of row and column subsets to be chosen ($|I|$, $|J|$) for each rank parameter
\Statex \textbf{Output:} Similarity matrix $\Xi\in\R^{n\times n}$ for $X$
\For{$r=r_{\min}:r_{\max}$}
\For{$i=1:N$}
\State Draw $I,J$ from $[m],[n]$ with (or without) replacement from probability distributions $\{p^{(r)}_i\}$ and $\{q^{(r)}_j\}$, respectively
\State $U=X(I,J)$, $R=X(I,:)$
\State $Y = U^\dagger R$
\State Normalize columns of $Y$ in $\ell_2$
\State $\Xi_i = Y^TY$
\EndFor
\State\label{ITEM:median} $\Xi^{(r)} = \textnormal{abs}(\textnormal{median}(\Xi_1,\dots,\Xi_N))$
\State\label{ITEM:Weight} $\Xi^{(r)}= \texttt{SoftThreshold}(\Xi^{(r)})$
\State\label{ITEM:ClusterLabel} Cluster Labels$(r)=\texttt{SpectralClustering}(\Xi^{(r)})$
\EndFor
\State $r_{\textnormal{best}} = \displaystyle\argmin_r\; C(r)$ \textbackslash\textbackslash as in \eqref{EQN:Ncut}
\State \textbf{Return:} $\Xi^{(r_{\textnormal{best}})}$
\end{algorithmic}
\end{algorithm}

\subsection{Algorithm Details}

There are a few aspects of Algorithm \ref{ALG:RCUR} that merit further discussion. First, the cost function of a given clustering is defined as in \cite{rsimpaper} by

\begin{equation}\label{EQN:Ncut}
C(r) := \frac{\sum_{i=1}^k\textnormal{Cut}(A_i^{(r)})}{|\lambda_{r+1}(L)-\lambda_r(L)|},     
\end{equation}
where $\{A_i^{(r)}\}$ give the clustering from Line \ref{ITEM:ClusterLabel},  $\textnormal{Cut}(A):=\sum_{i\in A, j\in A^C} w_{ij}$, and $\lambda_i(L)$ are the eigenvalues of the random walk graph Laplacian $L:=I-D^{-1}\Xi$; here, $\Xi$ is considered to be the weight (adjacency) matrix of the graph implicitly described in Line \ref{ITEM:Weight}, and $D$ is its degree matrix ($D_{ii} = \sum_{j}\Xi_{ij}$).  The cost function \eqref{EQN:Ncut} indicates the best clustering as the one which yields small cut number as well as large spectral gap in the graph Laplacian; the latter indicates the likelihood that the clustering corresponds to approximately $r$ distinct connected components of the graph.

Note that the size of the columns and row subsets drawn (i.e., $|I|$ and $|J|$) may be taken to depend upon the rank parameter $r$ in the for loop. This allows one to take into account the theoretical sampling orders established in the CUR decomposition literature.  For instance, it is known that sampling $O(r\log r)$ columns and rows according to leverage score distributions of a rank $r$ matrix $X$ is sufficient to guarantee $X=CU^\dagger R$ with high probability \cite{yang2015explicit}. However, since those results give only asymptotic sampling orders (the constants in the big-O notation are unknown), we don't put too much emphasis on the sampling order. Instead, in our implementation, we select $\kappa r$ columns and/or rows for a prescribed $\kappa$. Numerical experiments in the sequel illustrate performace for a range of $\kappa$ values.

\begin{figure}[h]
\centering
\includegraphics[width=\textwidth]{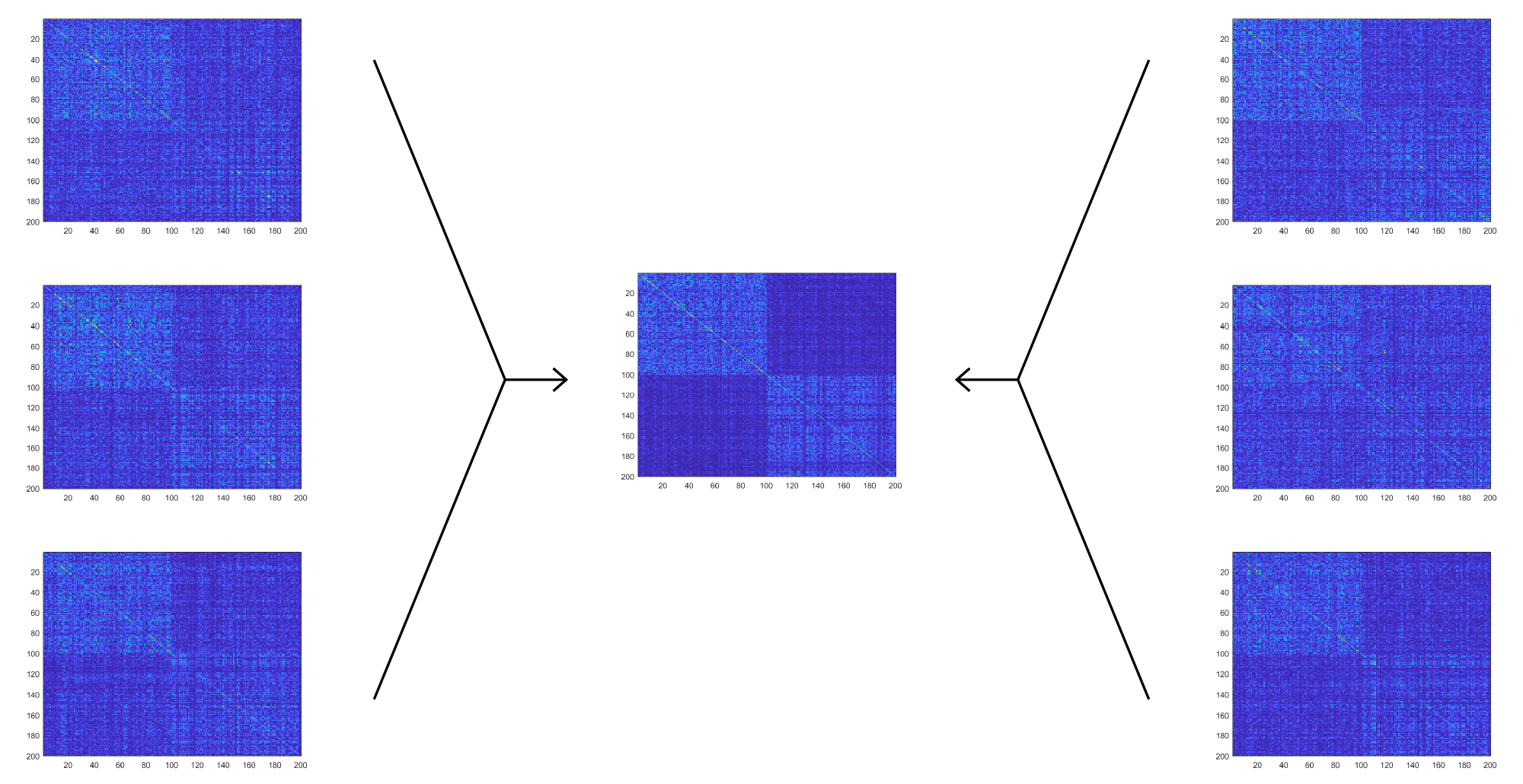}
\caption{Here, the data $X$ and noise $E$ are as in Figure \ref{FIG:ToyData}. The left and right column are noisy similarity matrices $\Xi_i$ as in Algorithm \ref{ALG:RCUR}. They are combined by taking entrywise median and soft thresholding to yield the cleaner similarity matrix $\Xi$ in the middle. }\label{FIG:RCUR}
\end{figure}

\paragraph{\textbf{E. Pluribus Unum}}  The idea behind Line \ref{ITEM:median} of Algorithm \ref{ALG:RCUR} is illustrated by a toy example in Figure \ref{FIG:RCUR}. Each $S_i$ may be a very noisy clustering matrix for $X$. However, aggregating them in some way (in this case we choose median for simplicity and robustness to outliers) can yield a candidate clustering matrix which is distinctly less noisy.

The choice of the \texttt{SoftThreshold} function can be specified by the user. We have not found this to have an overly significant effect on the performance of Algorithm \ref{ALG:RCUR} in thorough experimentation. We do however advocate for a skewed thresholding function, and in our implementation, we take the effect of $\texttt{SoftThreshold}(X)$ to be entrywise $X_{ij}^{\tau}$, where $\tau>1$ determines the behavior of the thresholding; this choice was based on the prior successes of \cite{rsimpaper,aldroubi2019cur}.

Since $\Xi^{(r)}$ is likely noisy due to the noise in the data (e.g., as in Figure \ref{FIG:ToyData}), we use Spectral Clustering to compute the final cluster labels.  In particular, we use Normalized Spectral Clustering which utilizes the symmetric normalized graph Laplacian with $\Xi^{(r)}$ as its weight matrix; that is, $L_{\textnormal{sym}}:=I-D^{-\frac12}\Xi^{(r)}D^{-\frac12}$ where $D$ is the corresponding degree matrix as defined above.  For further details, see \cite{von2007tutorial}, and note that normalized spectral clustering is also called normalized cuts, or NCut \cite{shi2000normalized}.

\section{Experimental Results}

In this section, we experimentally test the effect of using different sampling methods in Algorithm \ref{ALG:RCUR} when applied to real-world data. Motion segmentation and facial recognition are common applications for subspace clustering algorithms, and two benchmark datasets are Hopkins155 \cite{tron2007benchmark} and Yale Extended B \cite{yaledata}. In our experiments, we used the following sampling methods for columns and rows: Uniform (U), Leverage (Lev), Length (Len) and DEIM as described above, and selected $r$ rows and $\min(\texttt{\# Columns}, \kappa r)$ columns, where $\kappa$ is an oversampling parameter and $r$ is the rank estimation parameter in the outer for loop of Algorithm \ref{ALG:RCUR}. Informally, when we choose all columns, we say $\kappa=\infty$. In the experiments, we choose $\kappa\in \curly{1, 2, 5, \infty}$. 


For uniformity, every experiment for this paper was run on a 2016 Razer Blade 15 on Jupyter Notebooks with no GPU acceleration, enabling us to record the compute time per trial of each experiment. We note that experiments for this paper were done in Python, whereas those of \cite{aldroubi2019cur} and many previous works were done in Matlab. Implementation differences in eigenvalue solvers and packages such as spectral clustering will lead to minor variations in clustering performance between different implementations. Thus the error rates presented here may slightly vary from those in Matlab implementations. However, for a concrete comparison, we also ran the RSIM algorithm of \cite{rsimpaper} implemented in Python; RSIM typically yields the best performance on the given benchmarks, and is thus a faithful comparison for our method. We do not optimize all parameters here, but we execute the variants discussed herein consistently which allows for comparison of performance with respect to the parameters varied below. The Python code implementing our algorithm is available at \url{https://github.com/reeshadarian/RCUR
}.



\subsection{Motion Segmentation Data}

\begin{figure}[ht]
     \centering
     \begin{subfigure}[t]{0.3\textwidth}
         \centering
         \includegraphics[width=\textwidth]{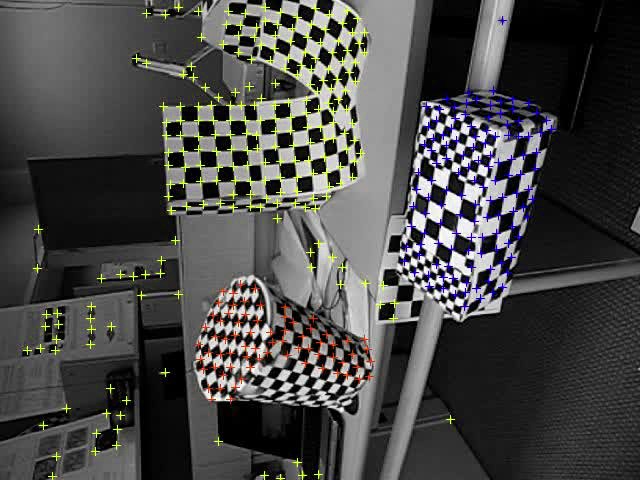}
     \end{subfigure}
     \begin{subfigure}[t]{0.3\textwidth}
         \centering
         \includegraphics[width=\textwidth]{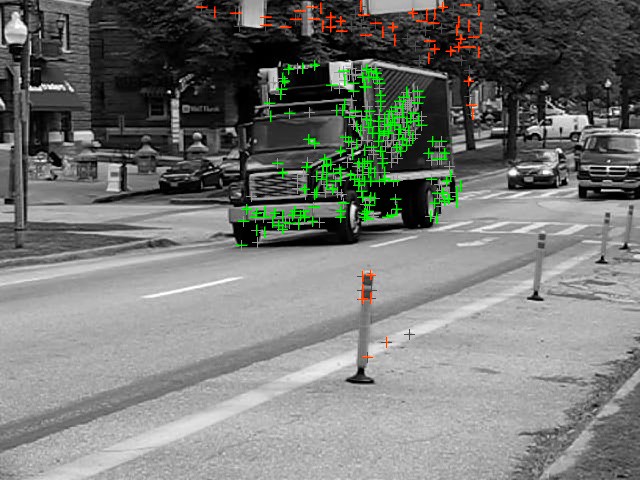}
     \end{subfigure}
     \begin{subfigure}[t]{0.3\textwidth}
         \centering
         \includegraphics[width=\textwidth]{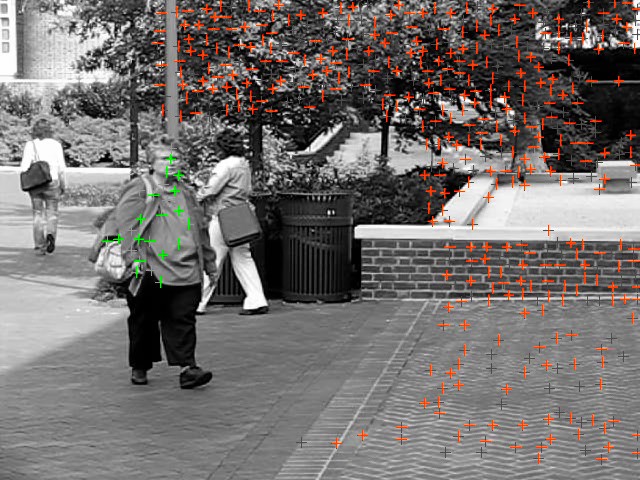}
     \end{subfigure}
        \caption{Preview image of a 2-motion (a) Checkerboard (b) Traffic (c) Articulated sequence from the Hopkins155 database}
        \label{fig:hopkins examples}
\end{figure}

Hopkins155 \cite{tron2007benchmark} is a commonly used dataset to test multi-body motion segmentation algorithms. The database consists of 155 videos of 39-550 motion-tracked points associated with an affine embedding which we use as the data matrix. The dataset is divided into checkerboard sequences, consisting of translations and rotations of rigid checkerboard patterns, traffic sequences, consisting of points belonging to motion-tracked vehicles, and articulated/non-rigid sequences, consisting of points belonging to human motion and other non-rigid objects such as cranes; see Figure \ref{fig:hopkins examples} for example still images from Hopkins155. Checkerboard sequences have the least amount of tracking noise, followed by traffic and articulated sequences.

Following \cite{rsimpaper}, we ran our algorithm on only the 2- and 3-motions sequences from each category. Since the labels were known \textit{a priori}, we were able to calculate percent clustering error. To account for the random variables in our algorithm, the mean of the errors from ten trials of each video sequence was considered. The percent error values in Tables \ref{tab:hop-rank-mult-all} and \ref{tab:hop-rank-mult-1}-\ref{tab:hop-rank-mult-5} are the averages of the mean errors pertaining to each type of sequence. For comparison we also ran the RSIM algorithm detailed in \cite{rsimpaper}. We refer our readers to \cite{aldroubi2019cur} for a comparison of RCUR with Uniform sampling and other popular subspace clustering algorithms on the Hopkins155 dataset.  In Table \ref{tab:hop-rank-mult-all} and \ref{tab:hop-rank-mult-1}-\ref{tab:hop-rank-mult-5}, the categories of Hopkins155 are of the form $X\#$ where the number 2 or 3 describes the number of motions tracked in the video sequence, and C = Checkerboard, T = Traffic, A = Articulated motion, and O = Overall (i.e., O2 is all 2-motion sequences from C2, T2, and A2) as described in \cite{tron2007benchmark}. Times in subsequent tables are per data matrix. 

\begin{table}[h]
\centering
\begin{tabular}{ccccccccccc}
\hline
\multicolumn{1}{|c|}{\multirow{2}{*}{\textbf{Category}}} & \multicolumn{5}{c|}{\textbf{Mean}} & \multicolumn{5}{c|}{\textbf{Median}} \\ \cline{2-11} 
\multicolumn{1}{|c|}{} & \textbf{U} & \textbf{Lev} & \textbf{Len} & \textbf{DEIM} & \multicolumn{1}{c|}{\textbf{RSIM}} & \textbf{U} & \textbf{Lev} & \textbf{Len} & \textbf{DEIM} & \multicolumn{1}{c|}{\textbf{RSIM}} \\ \cline{2-11} 
\multicolumn{1}{|c|}{C2} & \textbf{0.48} & 0.82 & 12.77 & 1.60 & \multicolumn{1}{c|}{1.02} & 0.00 & 0.00 & 7.64 & 0.00 & \multicolumn{1}{c|}{0.00} \\
\multicolumn{1}{|c|}{T2} & \textbf{0.66} & 1.90 & 4.70 & 1.36 & \multicolumn{1}{c|}{1.23} & 0.00 & 0.00 & 0.00 & 0.00 & \multicolumn{1}{c|}{0.00} \\
\multicolumn{1}{|c|}{O2} & \textbf{1.31} & 2.02 & 5.24 & 3.16 & \multicolumn{1}{c|}{1.43} & 0.00 & 0.00 & 0.00 & 0.00 & \multicolumn{1}{c|}{0.00} \\
\multicolumn{1}{|c|}{A2} & \textbf{0.61} & 1.21 & 9.99 & 1.68 & \multicolumn{1}{c|}{1.11} & 0.00 & 0.00 & 2.08 & 0.00 & \multicolumn{1}{c|}{0.00} \\
\multicolumn{1}{|c|}{C3} & 1.07 & \textbf{0.42} & 15.05 & 0.92 & \multicolumn{1}{c|}{0.62} & 0.37 & 0.15 & 9.92 & 0.41 & \multicolumn{1}{c|}{0.32} \\
\multicolumn{1}{|c|}{T3} & \textbf{0.87} & 5.61 & 3.15 & 4.08 & \multicolumn{1}{c|}{3.77} & 0.00 & 0.00 & 0.16 & 0.00 & \multicolumn{1}{c|}{0.19} \\
\multicolumn{1}{|c|}{A3} & \textbf{3.72} & 9.02 & \textbf{3.72} & 21.08 & \multicolumn{1}{c|}{18.95} & 3.72 & 9.02 & 3.72 & 21.08 & \multicolumn{1}{c|}{18.95} \\
\multicolumn{1}{|c|}{O3} & \textbf{1.18} & 1.95 & 12.02 & 2.70 & \multicolumn{1}{c|}{2.30} & 0.24 & 0.25 & 7.35 & 0.39 & \multicolumn{1}{c|}{0.37} \\ \hline
 &  &  &  &  &  &  &  &  &  &  \\ \hline
\multicolumn{1}{|c|}{\multirow{2}{*}{\textbf{Category}}} & \multicolumn{5}{c|}{\textbf{Standard Deviation}} & \multicolumn{5}{c|}{\textbf{Time (s)}} \\ \cline{2-11} 
\multicolumn{1}{|c|}{} & \textbf{U} & \textbf{Lev} & \textbf{Len} & \textbf{DEIM} & \multicolumn{1}{c|}{\textbf{RSIM}} & \textbf{U} & \textbf{Lev} & \textbf{Len} & \textbf{DEIM} & \multicolumn{1}{c|}{\textbf{RSIM}} \\ \cline{2-11} 
\multicolumn{1}{|c|}{C2} & 1.83 & 5.14 & 14.84 & 7.15 & \multicolumn{1}{c|}{5.45} & 1.14 & 1.34 & 1.54 & 0.37 & \multicolumn{1}{c|}{0.65} \\
\multicolumn{1}{|c|}{T2} & 3.25 & 6.95 & 9.90 & 4.46 & \multicolumn{1}{c|}{4.48} & 1.03 & 1.21 & 1.47 & 0.36 & \multicolumn{1}{c|}{0.65} \\
\multicolumn{1}{|c|}{A2} & 2.55 & 4.10 & 8.07 & 5.68 & \multicolumn{1}{c|}{3.04} & 0.68 & 0.84 & 0.98 & 0.24 & \multicolumn{1}{c|}{0.42} \\
\multicolumn{1}{|c|}{O2} & 2.36 & 5.61 & 13.74 & 6.45 & \multicolumn{1}{c|}{5.03} & 1.07 & 1.26 & 1.47 & 0.36 & \multicolumn{1}{c|}{0.63} \\
\multicolumn{1}{|c|}{C3} & 1.95 & 0.59 & 15.18 & 2.07 & \multicolumn{1}{c|}{0.78} & 2.88 & 3.27 & 3.76 & 1.16 & \multicolumn{1}{c|}{2.01} \\
\multicolumn{1}{|c|}{T3} & 1.92 & 9.73 & 5.57 & 9.48 & \multicolumn{1}{c|}{6.23} & 1.90 & 2.16 & 2.63 & 0.82 & \multicolumn{1}{c|}{1.42} \\
\multicolumn{1}{|c|}{A3} & 3.52 & 8.75 & 3.72 & 20.41 & \multicolumn{1}{c|}{18.28} & 0.28 & 0.38 & 0.45 & 0.16 & \multicolumn{1}{c|}{0.25} \\
\multicolumn{1}{|c|}{O3} & 2.16 & 5.55 & 14.31 & 8.19 & \multicolumn{1}{c|}{6.76} & 2.54 & 2.88 & 3.34 & 1.03 & \multicolumn{1}{c|}{1.79} \\ \hline

\end{tabular}
\caption{\% Classification Error on Hopkins155 for Algorithm \ref{ALG:RCUR} with $\kappa = \infty$ and various sampling methods (U = Uniform, Lev = Leverage Score, Len = Column/Row Length, and DEIM) and RSIM for comparison (bolded values are best of each category).}
\label{tab:hop-rank-mult-all}
\end{table}

For most of the Hopkins155 data matrices, the rank $r$ is essentially known from the analysis of Costeira and Kanade \cite{costeira1998multibody} to be at most 4 times the number of motions. In our experiments, we set $r_{\min}$ to be the number of motions, and $r_{\max}$ to be 4 times this number.  In Algorithm \ref{ALG:RCUR}, for each rank $r\in[r_{\min},r_{\max}]$, we use exactly $r$ rows selected from the probability distribution (or DEIM) specified in the trial, and we select $\kappa r$ columns according to the specified method in Line 3 of Algorithm \ref{ALG:RCUR}. The best overall results on Hopkins155 were obtained using Uniform sampling with $\kappa = \infty$, as shown in Table \ref{tab:hop-rank-mult-all}, though for some sequences (C3 and A3) Leverage score sampling and Length sampling performed the same or better on average.  This result essentially mimics the corresponding observations of \cite{aldroubi2019cur}. Results for other $\kappa$ values are shown in Appendix \ref{APP:Hopkins} for completeness. 

The good performance of using $\kappa = \infty$ comes at the expense of some extra computation time. The better performance may be attributed to the fact that utilizing all columns yields a noisy representation of the data in a frame dictionary with many elements, and the redundancy provided in the frame representation provides more robustness to noise. 

Note from Table \ref{tab:hop-rank-mult-all} that overall, Uniform sampling provides better accuracy, but this is not the case over all types of motion sequences. Indeed, Leverage score sampling achieves less error on the C3 sequences. The poor performance of Length sampling on Hopkins can be explained in the following way: the motions tracked in the video sequences actually result in affine subspaces rather than linear ones, and the matrix factorization methods are framed only for linear subspaces. However, to get around this issue, the points obtained from motion tracking are embedded in $\R^4$ in homogeneous coordinates. Consequently, Hopkins155 data matrices naturally have many all ones rows, which being distinctly larger in norm, are frequently chosen when sampling rows according to length sampling. We will see later that Length sampling is not always as bad as this for other datasets.

\begin{figure}[h]
     \centering
     \begin{subfigure}[t]{0.49\textwidth}
         \centering
         \includegraphics[width=\textwidth]{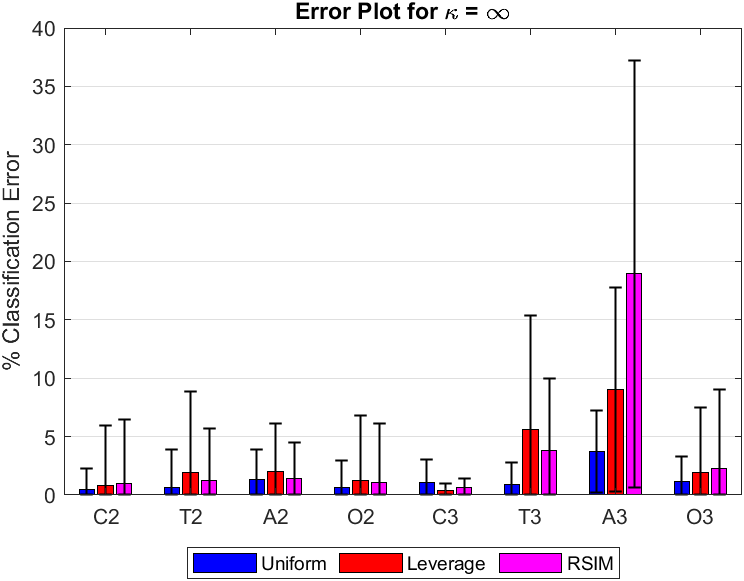}
         \caption{}
         \label{fig:hopkins kappa non-deim}
     \end{subfigure}
     \begin{subfigure}[t]{0.49\textwidth}
         \centering
         \includegraphics[width=\textwidth]{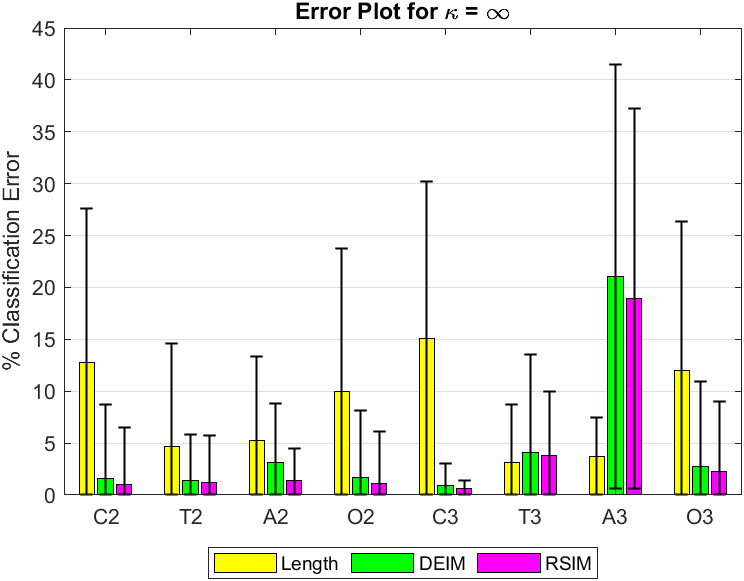}
         \caption{}
         \label{fig:hopkins kappa deim}
     \end{subfigure}
        \caption{Error Plots for Hopkins155 using Algorithm \ref{ALG:RCUR} with $\kappa = \infty$ for (a) Uniform and Leverage, and (b) Length and DEIM sampling. RSIM errors are plotted for comparison.}
        \label{fig:hopkins kappa infty}
\end{figure}

As can be seen from Figure \ref{fig:hopkins kappa non-deim}, RCUR with Uniform sampling typically performs better than the RSIM method, which on the whole performs a little better than the Leverage score sampling method. Deviation from this observation can be seen for the 3-motions Checkerboard and Articulated sequences. As noted before, the Checkerboard sequences are known to fit the union of subspaces model very well, which is demonstrated from the very low errors for all three methods. 
On the other hand, the Articulated sequences are known to fit the union of subspaces model poorly, which is reflected by the very high errors for all models. Even though Leverage score sampling performs much better than RSIM on average, it can be seen that the standard deviation for the former is rather large. The better performance in the mean errors is more likely due to the higher robustness to noise in the RCUR algorithm in general. Nevertheless, the main observation we draw is that Uniform sampling performs the best overall, while also taking the least time to run.  We offer two possible explanations for this: first that the columns and rows of Hopkins155 have very low incoherence.  That is, the quantities $\mu_1(X) = \sqrt{m/k}\max_{i}\|U_k(i,:)\|_2$ and $\mu_2(X) = \sqrt{n/k}\max_{i}\|V_k(i.:)\|_2$ are essentially constant; here $U_k$ (respectively, $V_k$) are the first $k$ left (respectively, right) singular vectors of $X$. 
An additional reason may be that the Leverage score distributions have a rank parameter in them, which requires a good estimate for the rank of the data matrix; for Hopkins the rank is typically known, but it could be that the noise is sufficient enough to counteract this when performing leverage score sampling. 

On the other hand, the Length and DEIM methods perform much poorer than RSIM, as can be seen from \ref{fig:hopkins kappa deim}. On the whole, DEIM sampling performs a little better than Length sampling. As noted previously, this appears to be a highly dataset-dependent phenomenon given that Hopkins155 matrices have all ones rows which hold no information. At worst, DEIM chooses one such row since any others will be discarded when the residuals are calculated in the DEIM algorithm. Length sampling does not guarantee this, and may choose many such rows, leading to significant error.

\begin{figure}
     \centering
     \begin{subfigure}[t]{0.49\textwidth}
         \centering
         \includegraphics[width=\textwidth]{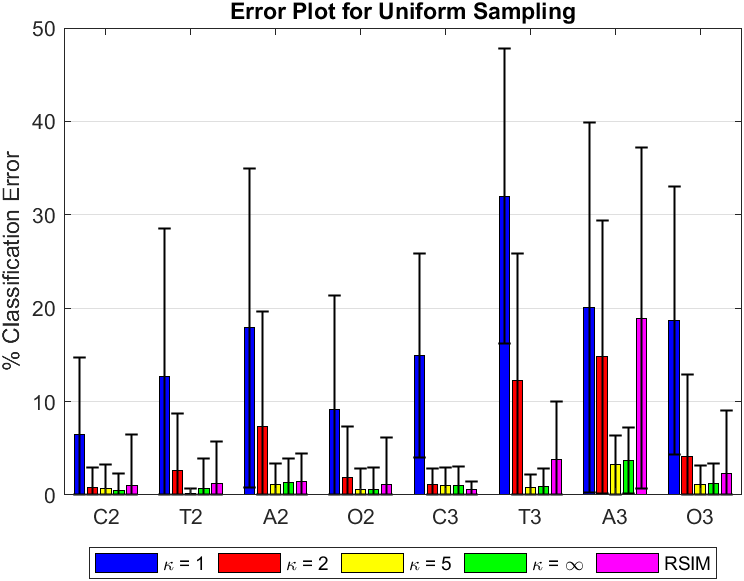}
         \caption{}
         \label{fig:hopkins uniform all}
     \end{subfigure}
     \begin{subfigure}[t]{0.49\textwidth}
         \centering
         \includegraphics[width=\textwidth]{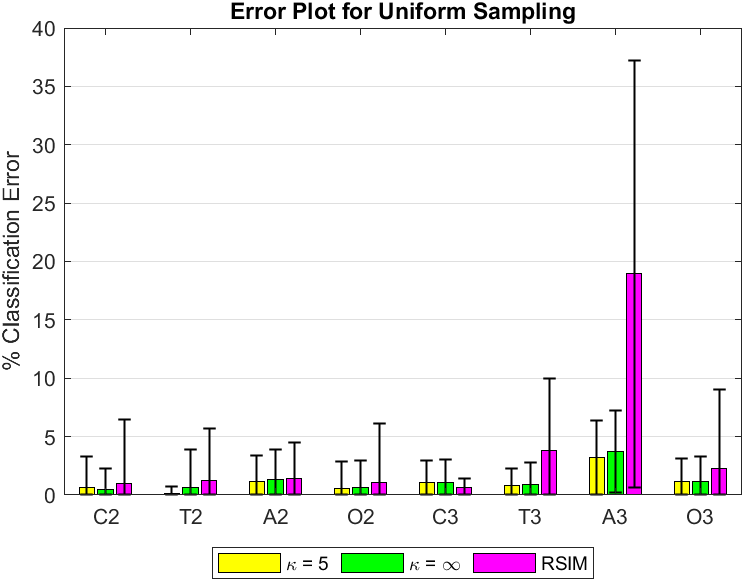}
         \caption{}
         \label{fig:hopkins uniform zoom}
     \end{subfigure}
        \caption{Error Plots for Hopkins155 using Uniform Sampling for (a) all $\kappa$ values (b) $\kappa\in \curly{5, \infty}$}
        \label{fig:hopkins uniform}
\end{figure}

Figure \ref{fig:hopkins uniform} shows the effect of using different values of $\kappa$ when performing Uniform sampling. As noted before, with more columns sampled, the redundancy of the representation increases, and the overall error goes down. It is evident from Figure \ref{fig:hopkins uniform zoom} than the gain in accuracy from the $\kappa = 5$ to $\kappa = \infty$ is relatively small. Unlike the Yale Extended B runs in the following section, the Hopkins155 data matrices are relatively small, so the difference in the number of columns actually selected for $\kappa=5$ and $\kappa=\infty$ may be negligible. This indicates that some consideration needs to be paid to the actual structure of the dataset when picking good values for $\kappa$.

\subsection{Facial Recognition}

\begin{figure}[h!]
     \centering
     \begin{subfigure}[t]{0.3\textwidth}
         \centering
         \includegraphics[width=\textwidth]{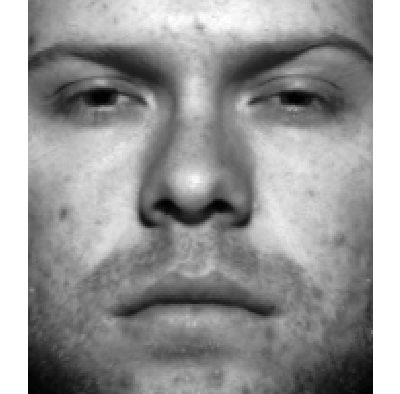}
     \end{subfigure}
     \begin{subfigure}[t]{0.3\textwidth}
         \centering
         \includegraphics[width=\textwidth]{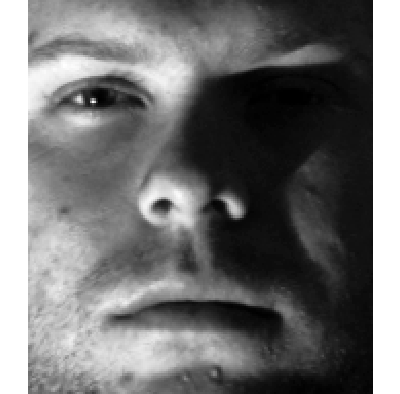}
     \end{subfigure}
        \caption{Sample images of subject 1 from the Yale Extended B database \cite{yaledata}.}
        \label{fig:yale examples}
\end{figure}

\begin{figure}
     \centering
     \begin{subfigure}[t]{0.49\textwidth}
         \centering
         \includegraphics[width=\textwidth]{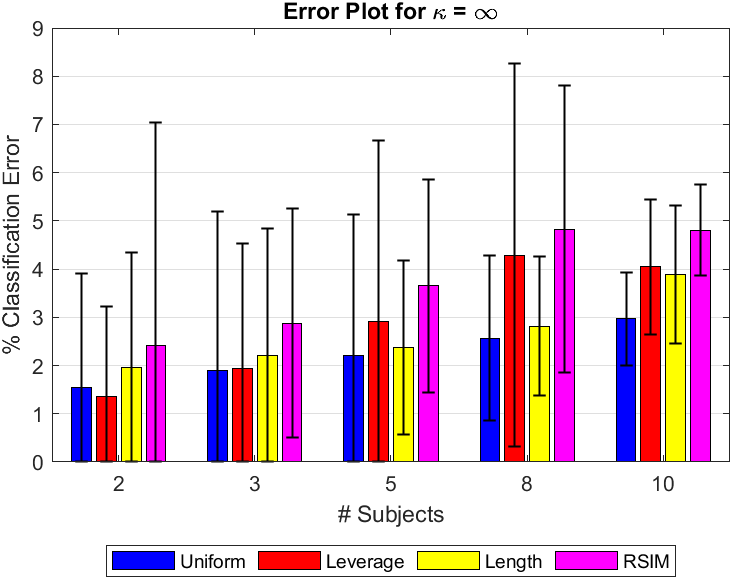}
         \caption{}
         \label{fig:yale kappa non-deim}
     \end{subfigure}
     \begin{subfigure}[t]{0.49\textwidth}
         \centering
         \includegraphics[width=\textwidth]{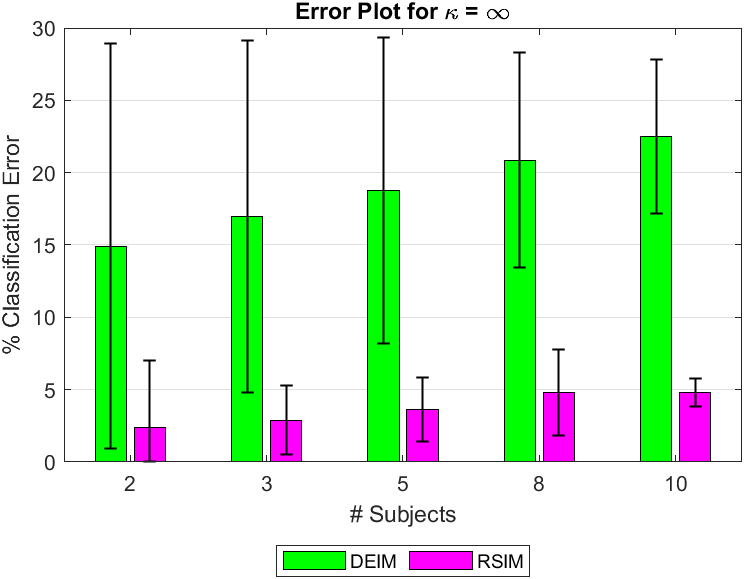}
         \caption{}
         \label{fig:yale kappa deim}
     \end{subfigure}
        \caption{Error Plots for Yale Extended B using $\kappa = \infty$ for (a) Uniform, Leverage, and Length Sampling, and (b) DEIM sampling}
        \label{fig:yale kappa infty}
\end{figure}

\begin{figure}
     \centering
     \begin{subfigure}[t]{0.49\textwidth}
         \centering
         \includegraphics[width=\textwidth]{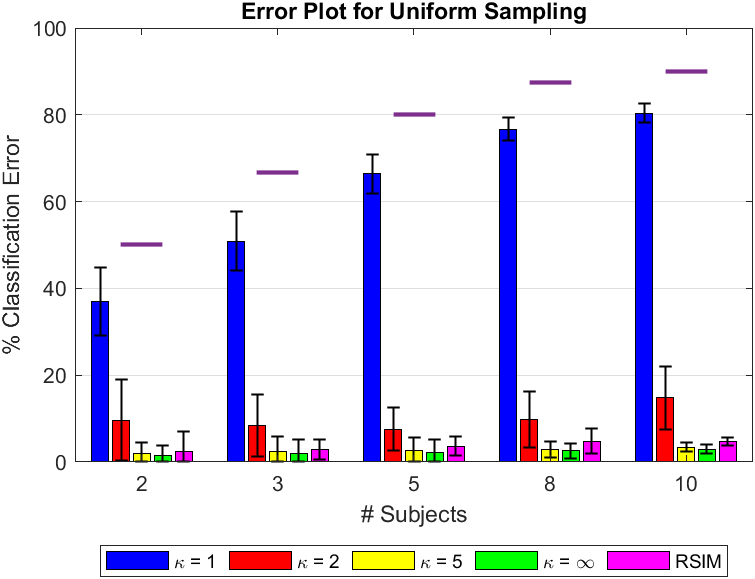}
         \caption{}
         \label{fig:yale uniform all}
     \end{subfigure}
     \begin{subfigure}[t]{0.49\textwidth}
         \centering
         \includegraphics[width=\textwidth]{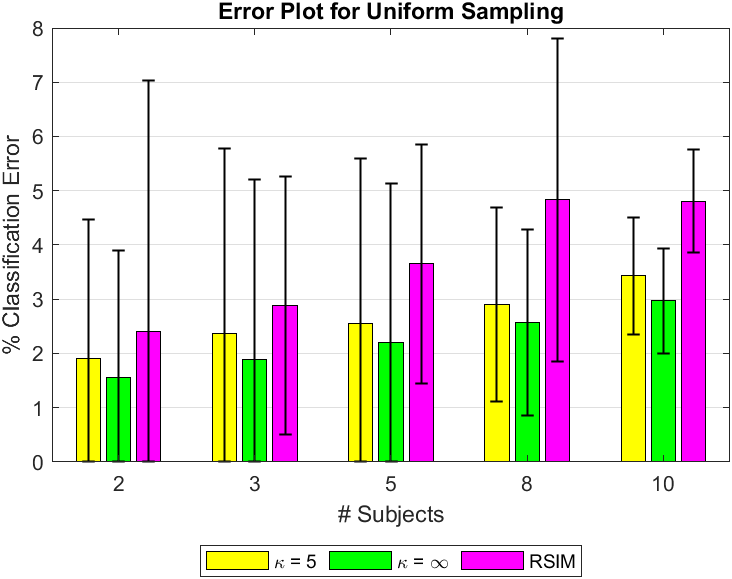}
         \caption{}
         \label{fig:yale uniform zoom}
     \end{subfigure}
        \caption{Error Plots for Yale Extended B using Uniform Sampling for (a) all $\kappa$ values (horizontal bar represents the maximum possible error) (b) $\kappa\in \curly{5, \infty}$}
        \label{fig:yale uniform}
\end{figure}

According to \cite{lambertianreflectance}, facial images illuminated from different directions can be described approximately via a union of subspaces model with images of one face being well-approximated via a 9-dimensional subspace. To this end, we ran the RCUR algorithm on the Yale Extended B facial database \cite{yaledata} containing photos of 38 subjects, each under 64 lighting conditions (see Figure \ref{fig:yale examples} for example images). It is to be noted that since the dataset was heavily corrupted, some data points indicated by the creators of the database were discarded prior to experimentation. Following the procedure of \cite{sparsesubspaceclustering}, we divide the subjects into four groups (subjects 1 to 10, 11 to 20, 21 to 30, and 31 to 38). Within each group, we ran our experiments on all combinations of 2, 3, 5, 8, and 10 (where possible) subjects and aggregated the errors to report final statistics for each number of subjects. Due to the large number of combinations of 5 subjects, only 500 combinations were randomly sampled without replacement. Finally, to test the stability of RCUR for higher dimensions and number of subspaces, we ran the experiment on the entire dataset of 38 subjects. For comparison, we also ran the RSIM algorithm detailed in \cite{rsimpaper} for all number of subjects except for the entire 38 person dataset.  The latter omission is due to the prohibitive time required to compute the SVD of the matrix arising from the full Yale dataset.  Note that no statistics are presented in Table \ref{tab:yale-rank-mult-all} for 38 subjects as there is only one data matrix for this number of subjects (as opposed to there being many combinations of 2 subjects, for example). We also note that the computation time for the trial on 38 subjects is error-prone, and thus is not reported. All results pertinent to this section can be found in Tables \ref{tab:yale-rank-mult-all} and \ref{tab:rankmult1}-\ref{tab:rankmult5}, and we aggregate the essential performance information in Figures \ref{fig:yale kappa infty} and \ref{fig:yale uniform}.
 
For each trial in each experiment, the $192\times 168$ pixel images were flattened into vectors of length 32256 and concatenated into a matrix whose columns correspond to facial images. As noted previously, the RCUR algorithm performs CUR decompositions based on a range of rank guesses; for the Yale data matrices, we estimate $r_{\min}$ and $r_{\max}$ to be 6 times the number of subjects and 10 times the number of subjects, respectively. We again used the Uniform, Leverage, Length and DEIM samplings, selecting a $r$ rows and $\min(\texttt{\# Columns}, \kappa r)$, where $\kappa\in \curly{1, 2, 5, \infty}$ is a parameter. 


Similar to the experiments on Hopkins155, our best results for Yale were found using Uniform sampling with $\kappa = \infty$ shown in Table \ref{tab:yale-rank-mult-all}. We again single out two illuminating sets of results. Among the values of $\kappa$, $\kappa = \infty$ was the best overall. 
We differentiate the DEIM method since it performed considerably poorer than the rest of the sampling methods. As can be seen from Figure \ref{fig:yale kappa non-deim}, RCUR with Uniform, Leverage, and Length sampling all perform better than the RSIM method. Among these, even though leverage score sampling performs well on 2- and 3-subject runs, it under-performs on the larger datasets. Uniform sampling, on the other hand, performs the best overall, while also taking the least time to run. These observations are in concert with the fact that the Yale Extended B data matrices also very low column-wise incoherence.  On the other hand, DEIM sampling performs very poorly overall, as can be seen on Figure \ref{fig:yale kappa deim}.  We find this surprising given that DEIM takes into account the singular vector information in a manner similar in spirit to leverage score sampling.  It is possible that the poor performance of DEIM is due to poor rank estimation or inexactness of the union of subspaces model. In Figure \ref{fig:yale kappa deim}, we include the deterministic DEIM sampling to choose rows, and use all columns (i.e., $\kappa=\infty$). This sampling method performs much worse than most other subspace clustering methods on Yale, which is somewhat surprising. However, we note that using the shape interaction matrix based purely on the SVD of the data also performs poorly overall, so it may simply be that DEIM is too rigid to account for the effects of noise.

It should be noted that for smaller values of $\kappa$, Leverage score sampling may perform better than Uniform sampling; see Appendix \ref{APP:Yale}. This is not unexpected given that Leverage scores use more information about the data matrix. However, the time required to run Algorithm \ref{ALG:RCUR} with Leverage scores for small $\kappa$ is larger than the time required to run Algorithm \ref{ALG:RCUR} with Uniform sampling for $\kappa=\infty$, so the benefit in performance of the former is mitigated by the time required. Table \ref{tab:yale-comparison} shows a comparison of RCUR with Uniform sampling and $\kappa=\infty$ with the Shape Interaction Matrix (SIM) \cite{costeira1998multibody}, Sparse Subspace Clustering (SSC) \cite{elhamifar2011sparse, elhamifar2013sparse}, Low-Rank Representation (LRR and LRR-H) \cite{liu2012robust}, Efficient Dense Subspace Clustering (EDSC and EDSC-H) \cite{ji2014efficient}, and RSIM. In general, RCUR performs significantly better overall.

\begin{table}[ht]
\centering
\begin{tabular}{ccccccccccc}
\hline
\multicolumn{1}{|c|}{\multirow{2}{*}{\textbf{Subjects}}} & \multicolumn{5}{c|}{\textbf{Mean}} & \multicolumn{5}{c|}{\textbf{Median}} \\ \cline{2-11} 
\multicolumn{1}{|c|}{} & \textbf{U} & \textbf{Lev} & \textbf{Len} & \textbf{DEIM} & \multicolumn{1}{c|}{\textbf{RSIM}} & \textbf{U} & \textbf{Lev} & \textbf{Len} & \textbf{DEIM} & \multicolumn{1}{c|}{\textbf{RSIM}} \\ \cline{2-11} 

\multicolumn{1}{|c|}{2} & 1.55 & \textbf{1.35} & 1.95 & 14.91 & \multicolumn{1}{c|}{2.41} & 0.78 & 0.78 & 1.56 & 9.38 & \multicolumn{1}{c|}{0.82} \\
\multicolumn{1}{|c|}{3} & \textbf{1.89} & 1.93 & 2.20 & 16.95 & \multicolumn{1}{c|}{2.88} & 1.06 & 1.08 & 1.57 & 12.80 & \multicolumn{1}{c|}{2.60} \\
\multicolumn{1}{|c|}{5} & \textbf{2.21} & 2.92 & 2.37 & 18.76 & \multicolumn{1}{c|}{3.65} & 1.60 & 2.22 & 2.19 & 15.19 & \multicolumn{1}{c|}{3.44} \\
\multicolumn{1}{|c|}{8} & \textbf{2.57} & 4.29 & 2.82 & 20.87 & \multicolumn{1}{c|}{4.83} & 2.36 & 3.52 & 2.93 & 20.28 & \multicolumn{1}{c|}{4.33} \\
\multicolumn{1}{|c|}{10} & \textbf{2.97} & 4.05 & 3.89 & 22.49 & \multicolumn{1}{c|}{4.81} & 3.46 & 3.91 & 3.59 & 25.16 & \multicolumn{1}{c|}{5.35} \\
\multicolumn{1}{|c|}{38} & \textbf{6.50} & 6.84 & 7.13 & 32.19 & \multicolumn{1}{c|}{-} & - & - & - & - & \multicolumn{1}{c|}{-} \\ \hline
 &  &  &  &  &  &  &  &  &  &  \\ \hline
\multicolumn{1}{|c|}{\multirow{2}{*}{\textbf{Subjects}}} & \multicolumn{5}{c|}{\textbf{Standard Deviation}} & \multicolumn{5}{c|}{\textbf{Time (s)}} \\ \cline{2-11} 
\multicolumn{1}{|c|}{} & \textbf{U} & \textbf{Lev} & \textbf{Len} & \textbf{DEIM} & \multicolumn{1}{c|}{\textbf{RSIM}} & \textbf{U} & \textbf{Lev} & \textbf{Len} & \textbf{DEIM} & \multicolumn{1}{c|}{\textbf{RSIM}} \\ \cline{2-11} 
\multicolumn{1}{|c|}{2} & 2.35 & 1.88 & 2.40 & 13.99 & \multicolumn{1}{c|}{4.63} & 6.30 & 14.03 & 6.57 & 8.05 & \multicolumn{1}{c|}{3.66} \\
\multicolumn{1}{|c|}{3} & 3.31 & 2.61 & 2.65 & 12.18 & \multicolumn{1}{c|}{2.38} & 37.60 & 81.63 & 38.17 & 50.22 & \multicolumn{1}{c|}{8.02} \\
\multicolumn{1}{|c|}{5} & 2.92 & 3.74 & 1.81 & 10.56 & \multicolumn{1}{c|}{2.21} & 127.82 & 271.95 & 125.72 & 163.25 & \multicolumn{1}{c|}{22.45} \\
\multicolumn{1}{|c|}{8} & 1.71 & 3.97 & 1.45 & 7.42 & \multicolumn{1}{c|}{2.98} & 224.22 & 402.82 & 208.23 & 243.02 & \multicolumn{1}{c|}{63.08} \\
\multicolumn{1}{|c|}{10} & 0.97 & 1.40 & 1.43 & 5.34 & \multicolumn{1}{c|}{0.95} & 17.30 & 32.75 & 17.70 & 19.50 & \multicolumn{1}{c|}{104.23} \\
\multicolumn{1}{|c|}{38} & - & - & - & - & \multicolumn{1}{c|}{-} & - & - & - & - & \multicolumn{1}{c|}{-} \\ \hline

\end{tabular}
\caption{\% Classification Error on Yale Extended B for Algorithm \ref{ALG:RCUR} with $\kappa = \infty$, and RSIM for comparison (bolded values are best of each category). }
\label{tab:yale-rank-mult-all}
\end{table}

\begin{table}[ht]
\centering
\begin{tabular}{cccccccccc}
\hline
\multicolumn{2}{c}{\textbf{Methods}} & \textbf{SIM} & \textbf{SSC} & \textbf{LRR} & \textbf{LRR-H} & \textbf{EDSC} & \textbf{EDSC-H} & \textbf{RSIM} & \textbf{\begin{tabular}[c]{@{}c@{}}RCUR\\ (U, $\kappa = \infty$)\end{tabular}} \\ \hline

\multirow{2}{*}{\textbf{2-Subjects}} & \multicolumn{1}{c|}{Average} & 8.10 & 1.86 & 9.52 & 2.54 & 5.42 & 2.65 & 1.99 & \textbf{1.55} \\
 & \multicolumn{1}{c|}{Median} & 6.25 & \textbf{0.00} & 5.47 & 0.78 & 4.69 & 1.56 & 0.79 & 0.78 \\ \hline
\multirow{2}{*}{\textbf{3-Subjects}} & \multicolumn{1}{c|}{Average} & 24.64 & 3.10 & 19.52 & 4.21 & 14.05 & 3.86 & 2.88 & \textbf{1.89} \\
 & \multicolumn{1}{c|}{Median} & 16.67 & \textbf{1.04} & 14.58 & 2.60 & 8.33 & 3.13 & 2.60 & 1.06 \\ \hline
\multirow{2}{*}{\textbf{5-Subjects}} & \multicolumn{1}{c|}{Average} & 45.62 & 4.31 & 34.16 & 6.90 & 36.99 & 5.11 & 3.72 & \textbf{2.21} \\
 & \multicolumn{1}{c|}{Median} & 48.13 & 2.50 & 35.00 & 5.63 & 30.63 & 3.75 & 3.48 & \textbf{1.60} \\ \hline
\multirow{2}{*}{\textbf{8-Subjects}} & \multicolumn{1}{c|}{Average} & 57.05 & 5.85 & 41.19 & 14.34 & 54.24 & 6.07 & 4.87 & \textbf{2.57} \\
 & \multicolumn{1}{c|}{Median} & 55.96 & 4.49 & 43.75 & 14.34 & 48.73 & 4.88 & 4.53 & \textbf{2.36} \\ \hline
\multirow{2}{*}{\textbf{10-Subjects}} & \multicolumn{1}{c|}{Average} & 65.10 & 10.94 & 38.85 & 22.92 & 59.58 & 7.24 & 4.93 & \textbf{2.97} \\
 & \multicolumn{1}{c|}{Median} & 64.06 & 5.63 & 41.09 & 23.59 & 50.47 & 6.09 & 5.16 & \textbf{3.46} \\ \hline

\end{tabular}
\caption{\% Classification Errors on Yale Extended B using various algorithms as well as Algorithm \ref{ALG:RCUR} with Uniform sampling and $\kappa=\infty$ (bolded values are best of each category).}
\label{tab:yale-comparison}
\end{table}

Our second set of results shown in Figure \ref{fig:yale uniform} correspond to Uniform sampling. As before, when $\kappa$ increases, the classification error goes down. Additionally, choosing $\kappa=2$ yields a markedly improved performance, which indicates that even small oversampling factors can yield meaningful labels, and are likely to perform better if the dataset is not as noisy as Yale Extended B (cf. similar observations from the Hopkins155 trials in Figure \ref{fig:hopkins uniform all} in which $\kappa=2$ yields very good performance on checkerboard sequences). One advantage of using small values of $\kappa$ is that the process is fast and memory-efficient, and can viably be used if resources are limited or data matrices are massive. One other noteworthy point is that as can be seen from the zoomed plot on Figure \ref{fig:yale uniform zoom}, even though $\kappa = \infty$ performs the best overall, the margin is not very large over choosing $\kappa = 5$, but the time for calculation approximately doubles. As a result, when resources are limited, using modest $\kappa$ is a viable compromise for some datasets. 

One of the most interesting takeaways from Table \ref{tab:yale-rank-mult-all} is that the random sampling variants of RCUR appear to scale quite well to higher-dimensional problems, as evidenced by the fact that the classification error for Uniform sampling grows slowly with respect to the number of subjects when compared with that of RSIM. Scalability of CUR decompositions has been explored by many works in the past, and is one of its primary benefits to computational applications.

\section{Conclusion}

The purpose of this paper was to better understand how various hyperparameters of the robust CUR similarity matrix algorithm (Algorithm \ref{ALG:RCUR}) effect clustering performance on different datasets. To understand these better, we tested primarily on Hopkins155 and Yale Extended B data. We tested various column and row sampling methods including randomly sampling with replacement according to Uniform, Length, and Leverage score distributions, and deterministic sampling via DEIM.  In the real applications, DEIM performed more poorly than would be expected given the good theoretical results in \cite{sorensen2016deim}.  On the other hand, the performance of random sampling appears to be somewhat more dependent on the structure of the data and the magnitude of the noise.  

In our trials, Uniform sampling often performed better, but Leverage score sampling performed similarly or better for some types of data, and at times exhibited smaller standard deviation on certain groups of data.  Consequently, it appears that typically Uniform sampling is adequate for forming CUR-based similarity matrices for subspace clustering applications, and given the ease of computation compared with Leverage score sampling, this is good news for practitioners.  One possible explanation for the success of Uniform sampling is that the data matrices involved appear to have constant incoherence compared to the rank and problem size.  

However, incoherence of data is not enough to guarantee success of clustering; the model must be accurate as well.  Following some other works, we tested Algorithm \ref{ALG:RCUR} on the MNIST handwritted digits dataset, but the performance was poor (20\% error on average).  While this performance is comparable and often better than other subspace clustering methods applied to MNIST, it suggests that the dataset (which has very small $\mu_1$ and $\mu_2$ even for large matrices) does not fit the union of subspaces model. Indeed, one hypothesis is that it exhibits more of a union of nonlinear manifold structure, which is known to be a much more complicated problem (e.g., \cite{elhamifar2011sparse}).  It would be interesting in the future to better understand the role of incoherence with respect to the success or failure of various subspace clustering algorithms.

As far as the other parameters we tested, we found that typically using all or a substantial proportion of the columns provides sufficient redundancy in the CUR representation to allow for good clustering performance. However, for larger data, it appears that modest values of the oversampling rate $\kappa$ can achieve good results.



\section*{Acknowledgments}

Both authors were supported in part by the National Science Foundation TRIPODS program, grant number CCF--1740858. R.A. was supported by REU funding through the UA--TRIPODS program.

Both authors were sponsored in part by the Army Research Office under grant number W911NF-20-1-0076. The views and conclusions contained in this document are those of the authors and should not be interpreted as representing the official policies, either expressed or implied, of the Army Research Office or the U.S. Government. The U.S. Government is authorized to reproduce and distribute reprints for Government purposes notwithstanding any copyright notation herein.

\bibliographystyle{plain}
\bibliography{references}

\begin{thebibliography}{10}

\bibitem{aldroubi2019cur}
Akram Aldroubi, Keaton Hamm, Ahmet~Bugra Koku, and Ali Sekmen.
\newblock Cur decompositions, similarity matrices, and subspace clustering.
\newblock {\em Frontiers in Applied Mathematics and Statistics}, 4:65, 2019.

\bibitem{aldroubi2012nearness}
Akram Aldroubi and Ali Sekmen.
\newblock Nearness to local subspace algorithm for subspace and motion
  segmentation.
\newblock {\em IEEE Signal Processing Letters}, 19(10):704--707, 2012.

\bibitem{aldroubi2018similarity}
Akram Aldroubi, Ali Sekmen, Ahmet~Bugra Koku, and Ahmet~Faruk Cakmak.
\newblock Similarity matrix framework for data from union of subspaces.
\newblock {\em Applied and Computational Harmonic Analysis}, 45(2):425--435,
  2018.

\bibitem{lambertianreflectance}
Ronen Basri and David~W Jacobs.
\newblock Lambertian reflectance and linear subspaces.
\newblock {\em IEEE transactions on pattern analysis and machine intelligence},
  25(2):218--233, 2003.

\bibitem{chiu2013sublinear}
Jiawei Chiu and Laurent Demanet.
\newblock Sublinear randomized algorithms for skeleton decompositions.
\newblock {\em SIAM Journal on Matrix Analysis and Applications},
  34(3):1361--1383, 2013.

\bibitem{costeira1998multibody}
Joao~Paulo Costeira and Takeo Kanade.
\newblock A multibody factorization method for independently moving objects.
\newblock {\em International Journal of Computer Vision}, 29(3):159--179, 1998.

\bibitem{donoho2006compressed}
David~L Donoho.
\newblock Compressed sensing.
\newblock {\em IEEE Transactions on information theory}, 52(4):1289--1306,
  2006.

\bibitem{sparsesubspaceclustering}
Ehsan Elhamifar and Ren{\'e} Vidal.
\newblock Sparse subspace clustering.
\newblock In {\em 2009 IEEE Conference on Computer Vision and Pattern
  Recognition}, pages 2790--2797. IEEE, 2009.

\bibitem{elhamifar2011sparse}
Ehsan Elhamifar and Ren{\'e} Vidal.
\newblock Sparse manifold clustering and embedding.
\newblock {\em Advances in neural information processing systems}, 24:55--63,
  2011.

\bibitem{elhamifar2013sparse}
Ehsan Elhamifar and Ren{\'e} Vidal.
\newblock Sparse subspace clustering: Algorithm, theory, and applications.
\newblock {\em IEEE transactions on pattern analysis and machine intelligence},
  35(11):2765--2781, 2013.

\bibitem{fefferman2016testing}
Charles Fefferman, Sanjoy Mitter, and Hariharan Narayanan.
\newblock Testing the manifold hypothesis.
\newblock {\em Journal of the American Mathematical Society}, 29(4):983--1049,
  2016.

\bibitem{foucart2013invitation}
Simon Foucart and Holger Rauhut.
\newblock An invitation to compressive sensing.
\newblock In {\em A mathematical introduction to compressive sensing}, pages
  1--39. Springer, 2013.

\bibitem{yaledata}
A.S. Georghiades, P.N. Belhumeur, and D.J. Kriegman.
\newblock From few to many: Illumination cone models for face recognition under
  variable lighting and pose.
\newblock {\em IEEE Trans. Pattern Anal. Mach. Intelligence}, 23(6):643--660,
  2001.

\bibitem{gu1996efficient}
Ming Gu and Stanley~C Eisenstat.
\newblock Efficient algorithms for computing a strong rank-revealing qr
  factorization.
\newblock {\em SIAM Journal on Scientific Computing}, 17(4):848--869, 1996.

\bibitem{hamm2020perspectives}
Keaton Hamm and Longxiu Huang.
\newblock Perspectives on cur decompositions.
\newblock {\em Applied and Computational Harmonic Analysis}, 48(3):1088--1099,
  2020.

\bibitem{ji2014efficient}
Pan Ji, Mathieu Salzmann, and Hongdong Li.
\newblock Efficient dense subspace clustering.
\newblock In {\em IEEE Winter Conference on Applications of Computer Vision},
  pages 461--468. IEEE, 2014.

\bibitem{rsimpaper}
Pan Ji, Mathieu Salzmann, and Hongdong Li.
\newblock Shape interaction matrix revisited and robustified: Efficient
  subspace clustering with corrupted and incomplete data.
\newblock In {\em Proceedings of the IEEE International Conference on computer
  Vision}, pages 4687--4695, 2015.

\bibitem{liu2012robust}
Guangcan Liu, Zhouchen Lin, Shuicheng Yan, Ju~Sun, Yong Yu, and Yi~Ma.
\newblock Robust recovery of subspace structures by low-rank representation.
\newblock {\em IEEE transactions on pattern analysis and machine intelligence},
  35(1):171--184, 2012.

\bibitem{liu2010robust}
Guangcan Liu, Zhouchen Lin, and Yong Yu.
\newblock Robust subspace segmentation by low-rank representation.
\newblock In {\em Proceedings of the 27th International Conference on
  International Conference on Machine Learning}, pages 663--670, 2010.

\bibitem{mahoney2009cur}
Michael~W Mahoney and Petros Drineas.
\newblock Cur matrix decompositions for improved data analysis.
\newblock {\em Proceedings of the National Academy of Sciences},
  106(3):697--702, 2009.

\bibitem{shi2000normalized}
Jianbo Shi and Jitendra Malik.
\newblock Normalized cuts and image segmentation.
\newblock {\em IEEE Transactions on pattern analysis and machine intelligence},
  22(8):888--905, 2000.

\bibitem{sorensen2016deim}
Danny~C Sorensen and Mark Embree.
\newblock A deim induced cur factorization.
\newblock {\em SIAM Journal on Scientific Computing}, 38(3):A1454--A1482, 2016.

\bibitem{szegedy2014intriguing}
Christian Szegedy, Wojciech Zaremba, Ilya Sutskever, Joan Bruna, Dumitru Erhan,
  Ian Goodfellow, and Rob Fergus.
\newblock Intriguing properties of neural networks.
\newblock In {\em 2nd International Conference on Learning Representations,
  ICLR 2014}, 2014.

\bibitem{tron2007benchmark}
Roberto Tron and Ren{\'e} Vidal.
\newblock A benchmark for the comparison of 3-d motion segmentation algorithms.
\newblock In {\em 2007 IEEE conference on computer vision and pattern
  recognition}, pages 1--8. IEEE, 2007.

\bibitem{vidal2011subspace}
Ren{\'e} Vidal.
\newblock Subspace clustering.
\newblock {\em IEEE Signal Processing Magazine}, 28(2):52--68, 2011.

\bibitem{von2007tutorial}
Ulrike Von~Luxburg.
\newblock A tutorial on spectral clustering.
\newblock {\em Statistics and computing}, 17(4):395--416, 2007.

\bibitem{yang2015explicit}
Tianbao Yang, Lijun Zhang, Rong Jin, and Shenghuo Zhu.
\newblock An explicit sampling dependent spectral error bound for column subset
  selection.
\newblock In {\em International Conference on Machine Learning}, pages
  135--143. PMLR, 2015.

\end{thebibliography}

\addresseshere

\newpage

\appendix
\section{Tables of other trials on Hopkins155}\label{APP:Hopkins}

\begin{table}[h]
\centering
\begin{tabular}{ccccccccc}
\hline
\multicolumn{1}{|c|}{\multirow{2}{*}{\textbf{Category}}} & \multicolumn{4}{c|}{\textbf{Mean}} & \multicolumn{4}{c|}{\textbf{Median}} \\ \cline{2-9} 
\multicolumn{1}{|c|}{} & \textbf{U} & \textbf{Lev} & \textbf{Len} & \multicolumn{1}{c|}{\textbf{DEIM}} & \textbf{U} & \textbf{Lev} & \textbf{Len} & \multicolumn{1}{c|}{\textbf{DEIM}} \\ \cline{2-9} 

\multicolumn{1}{|c|}{C2} & \textbf{6.50} & 8.54 & 17.70 & \multicolumn{1}{c|}{30.99} & 3.09 & 5.37 & 15.24 & \multicolumn{1}{c|}{30.85} \\
\multicolumn{1}{|c|}{T2} & 12.64 & \textbf{12.44} & 15.11 & \multicolumn{1}{c|}{22.40} & 2.24 & 7.77 & 10.29 & \multicolumn{1}{c|}{20.39} \\
\multicolumn{1}{|c|}{O2} & 17.89 & \textbf{16.23} & 18.91 & \multicolumn{1}{c|}{21.00} & 9.49 & 8.21 & 19.18 & \multicolumn{1}{c|}{17.71} \\
\multicolumn{1}{|c|}{A2} & \textbf{9.13} & 10.25 & 17.15 & \multicolumn{1}{c|}{27.85} & 3.34 & 6.56 & 14.00 & \multicolumn{1}{c|}{28.89} \\
\multicolumn{1}{|c|}{C3} & \textbf{14.97} & 20.45 & 18.11 & \multicolumn{1}{c|}{45.81} & 13.56 & 18.65 & 13.84 & \multicolumn{1}{c|}{47.05} \\
\multicolumn{1}{|c|}{T3} & 31.99 & 32.63 & \textbf{20.99} & \multicolumn{1}{c|}{30.93} & 27.97 & 30.68 & 9.67 & \multicolumn{1}{c|}{24.71} \\
\multicolumn{1}{|c|}{O3} & \textbf{20.08} & 18.17 & 20.11 & \multicolumn{1}{c|}{47.71} & 20.08 & 18.17 & 20.11 & \multicolumn{1}{c|}{47.71} \\
\multicolumn{1}{|c|}{A3} & \textbf{18.67} & 22.76 & 18.80 & \multicolumn{1}{c|}{42.94} & 14.59 & 20.53 & 13.38 & \multicolumn{1}{c|}{44.11} \\ \hline
\multicolumn{1}{l}{} & \multicolumn{1}{l}{} & \multicolumn{1}{l}{} & \multicolumn{1}{l}{} & \multicolumn{1}{l}{} & \multicolumn{1}{l}{} & \multicolumn{1}{l}{} & \multicolumn{1}{l}{} & \multicolumn{1}{l}{} \\ \hline
\multicolumn{1}{|c|}{\multirow{2}{*}{\textbf{Category}}} & \multicolumn{4}{c|}{\textbf{Standard Deviation}} & \multicolumn{4}{c|}{\textbf{Time (s)}} \\ \cline{2-9} 
\multicolumn{1}{|c|}{} & \textbf{U} & \textbf{Lev} & \textbf{Len} & \multicolumn{1}{c|}{\textbf{DEIM}} & \textbf{U} & \textbf{Lev} & \textbf{Len} & \multicolumn{1}{c|}{\textbf{DEIM}} \\ \cline{2-9} 
\multicolumn{1}{|c|}{C2} & 8.25 & 8.88 & 15.35 & \multicolumn{1}{c|}{7.34} & 0.77 & 0.78 & 0.77 & \multicolumn{1}{c|}{0.39} \\
\multicolumn{1}{|c|}{T2} & 15.84 & 12.83 & 15.76 & \multicolumn{1}{c|}{11.30} & 0.70 & 0.71 & 0.70 & \multicolumn{1}{c|}{0.37} \\
\multicolumn{1}{|c|}{O2} & 17.11 & 16.52 & 15.94 & \multicolumn{1}{c|}{9.33} & 0.46 & 0.47 & 0.46 & \multicolumn{1}{c|}{0.24} \\
\multicolumn{1}{|c|}{A2} & 12.27 & 11.19 & 15.57 & \multicolumn{1}{c|}{9.71} & 0.73 & 0.74 & 0.72 & \multicolumn{1}{c|}{0.37} \\
\multicolumn{1}{|c|}{C3} & 10.92 & 10.69 & 12.88 & \multicolumn{1}{c|}{6.58} & 2.45 & 2.47 & 2.50 & \multicolumn{1}{c|}{1.24} \\
\multicolumn{1}{|c|}{T3} & 15.77 & 13.12 & 14.96 & \multicolumn{1}{c|}{11.16} & 1.66 & 1.68 & 1.66 & \multicolumn{1}{c|}{0.85} \\
\multicolumn{1}{|c|}{O3} & 19.81 & 17.37 & 20.11 & \multicolumn{1}{c|}{7.82} & 0.25 & 0.30 & 0.28 & \multicolumn{1}{c|}{0.12} \\
\multicolumn{1}{|c|}{A3} & 14.37 & 12.70 & 13.87 & \multicolumn{1}{c|}{9.84} & 2.17 & 2.19 & 2.21 & \multicolumn{1}{c|}{1.10} \\ \hline
\end{tabular}
\caption{\% Classification Error on Hopkins155 for Algorithm \ref{ALG:RCUR} with $\kappa = 1$ under various sampling methods, (bolded values are best of each category).}
\label{tab:hop-rank-mult-1}
\end{table}

\begin{table}[h]
\centering
\begin{tabular}{ccccccccc}
\hline
\multicolumn{1}{|c|}{\multirow{2}{*}{\textbf{Category}}} & \multicolumn{4}{c|}{\textbf{Mean}} & \multicolumn{4}{c|}{\textbf{Median}} \\ \cline{2-9} 
\multicolumn{1}{|c|}{} & \textbf{U} & \textbf{Lev} & \textbf{Len} & \multicolumn{1}{c|}{\textbf{DEIM}} & \textbf{U} & \textbf{Lev} & \textbf{Len} & \multicolumn{1}{c|}{\textbf{DEIM}} \\ \cline{2-9} 

\multicolumn{1}{|c|}{C2} & \textbf{0.75} & 1.82 & 14.37 & \multicolumn{1}{c|}{9.23} & 0.00 & 0.00 & 10.35 & \multicolumn{1}{c|}{6.48} \\
\multicolumn{1}{|c|}{T2} & 2.60 & \textbf{2.28} & 9.11 & \multicolumn{1}{c|}{9.51} & 0.00 & 0.00 & 4.43 & \multicolumn{1}{c|}{9.37} \\
\multicolumn{1}{|c|}{O2} & 7.37 & \textbf{1.82} & 17.11 & \multicolumn{1}{c|}{9.06} & 0.78 & 0.13 & 15.34 & \multicolumn{1}{c|}{8.72} \\
\multicolumn{1}{|c|}{A2} & \textbf{1.83} & 1.94 & 13.26 & \multicolumn{1}{c|}{9.29} & 0.00 & 0.00 & 8.60 & \multicolumn{1}{c|}{7.37} \\
\multicolumn{1}{|c|}{C3} & \textbf{1.10} & 2.50 & 15.11 & \multicolumn{1}{c|}{18.75} & 0.52 & 0.76 & 11.03 & \multicolumn{1}{c|}{19.40} \\
\multicolumn{1}{|c|}{T3} & 12.29 & \textbf{8.29} & 13.10 & \multicolumn{1}{c|}{21.73} & 3.47 & 7.55 & 9.43 & \multicolumn{1}{c|}{22.57} \\
\multicolumn{1}{|c|}{O3} & 14.78 & 10.85 & \textbf{7.50} & \multicolumn{1}{c|}{19.33} & 14.78 & 10.85 & 7.50 & \multicolumn{1}{c|}{19.33} \\
\multicolumn{1}{|c|}{A3} & \textbf{4.12} & 4.13 & 14.27 & \multicolumn{1}{c|}{19.38} & 0.55 & 0.77 & 10.42 & \multicolumn{1}{c|}{22.01} \\ \hline
\multicolumn{1}{l}{} & \multicolumn{1}{l}{} & \multicolumn{1}{l}{} & \multicolumn{1}{l}{} & \multicolumn{1}{l}{} & \multicolumn{1}{l}{} & \multicolumn{1}{l}{} & \multicolumn{1}{l}{} & \multicolumn{1}{l}{} \\ \hline
\multicolumn{1}{|c|}{\multirow{2}{*}{\textbf{Category}}} & \multicolumn{4}{c|}{\textbf{Standard Deviation}} & \multicolumn{4}{c|}{\textbf{Time (s)}} \\ \cline{2-9} 
\multicolumn{1}{|c|}{} & \textbf{U} & \textbf{Lev} & \textbf{Len} & \multicolumn{1}{c|}{\textbf{DEIM}} & \textbf{U} & \textbf{Lev} & \textbf{Len} & \multicolumn{1}{c|}{\textbf{DEIM}} \\ \cline{2-9} 
\multicolumn{1}{|c|}{C2} & 2.18 & 6.29 & 14.35 & \multicolumn{1}{c|}{8.68} & 1.01 & 0.81 & 0.73 & \multicolumn{1}{c|}{0.37} \\
\multicolumn{1}{|c|}{T2} & 6.12 & 7.11 & 11.03 & \multicolumn{1}{c|}{7.11} & 0.92 & 0.72 & 0.67 & \multicolumn{1}{c|}{0.34} \\
\multicolumn{1}{|c|}{O2} & 12.31 & 3.56 & 14.22 & \multicolumn{1}{c|}{6.69} & 0.60 & 0.49 & 0.44 & \multicolumn{1}{c|}{0.23} \\
\multicolumn{1}{|c|}{A2} & 5.51 & 6.33 & 13.80 & \multicolumn{1}{c|}{8.13} & 0.95 & 0.76 & 0.69 & \multicolumn{1}{c|}{0.35} \\
\multicolumn{1}{|c|}{C3} & 1.71 & 5.35 & 14.69 & \multicolumn{1}{c|}{7.75} & 2.56 & 2.54 & 2.33 & \multicolumn{1}{c|}{1.11} \\
\multicolumn{1}{|c|}{T3} & 13.52 & 9.09 & 12.48 & \multicolumn{1}{c|}{4.31} & 1.74 & 1.70 & 1.59 & \multicolumn{1}{c|}{0.77} \\
\multicolumn{1}{|c|}{O3} & 14.58 & 10.85 & 7.50 & \multicolumn{1}{c|}{18.33} & 0.28 & 0.32 & 0.26 & \multicolumn{1}{c|}{0.14} \\
\multicolumn{1}{|c|}{A3} & 8.80 & 7.25 & 14.07 & \multicolumn{1}{c|}{8.30} & 2.26 & 2.25 & 2.06 & \multicolumn{1}{c|}{0.99} \\ \hline
\end{tabular}
\caption{\% Classification Error on Hopkins155 for Algorithm \ref{ALG:RCUR} with $\kappa = 2$ under various sampling methods, (bolded values are best of each category).}
\label{tab:hop-rank-mult-2}
\end{table}

\begin{table}[h]
\centering
\begin{tabular}{ccccccccc}
\hline
\multicolumn{1}{|c|}{\multirow{2}{*}{\textbf{Category}}} & \multicolumn{4}{c|}{\textbf{Mean}} & \multicolumn{4}{c|}{\textbf{Median}} \\ \cline{2-9} 
\multicolumn{1}{|c|}{} & \textbf{U} & \textbf{Lev} & \textbf{Len} & \multicolumn{1}{c|}{\textbf{DEIM}} & \textbf{U} & \textbf{Lev} & \textbf{Len} & \multicolumn{1}{c|}{\textbf{DEIM}} \\ \cline{2-9} 

\multicolumn{1}{|c|}{C2} & \textbf{0.67} & 1.48 & 14.40 & \multicolumn{1}{c|}{3.98} & 0.00 & 0.00 & 9.22 & \multicolumn{1}{c|}{0.19} \\
\multicolumn{1}{|c|}{T2} & \textbf{0.17} & 1.98 & 3.17 & \multicolumn{1}{c|}{3.34} & 0.00 & 0.00 & 0.09 & \multicolumn{1}{c|}{0.59} \\
\multicolumn{1}{|c|}{O2} & \textbf{1.16} & 2.11 & 8.11 & \multicolumn{1}{c|}{3.41} & 0.02 & 0.00 & 7.53 & \multicolumn{1}{c|}{0.95} \\
\multicolumn{1}{|c|}{A2} & \textbf{0.59} & 1.67 & 10.92 & \multicolumn{1}{c|}{3.76} & 0.00 & 0.00 & 6.00 & \multicolumn{1}{c|}{0.24} \\
\multicolumn{1}{|c|}{C3} & \textbf{1.06} & 1.33 & 15.46 & \multicolumn{1}{c|}{5.13} & 0.32 & 0.24 & 10.48 & \multicolumn{1}{c|}{2.18} \\
\multicolumn{1}{|c|}{T3} & \textbf{0.78} & 7.58 & 6.52 & \multicolumn{1}{c|}{8.40} & 0.09 & 0.00 & 5.39 & \multicolumn{1}{c|}{5.76} \\
\multicolumn{1}{|c|}{O3} & \textbf{3.22} & 3.81 & 9.38 & \multicolumn{1}{c|}{17.08} & 3.22 & 3.81 & 9.38 & \multicolumn{1}{c|}{17.08} \\
\multicolumn{1}{|c|}{A3} & \textbf{1.13} & 2.72 & 13.32 & \multicolumn{1}{c|}{6.47} & 0.26 & 0.21 & 9.24 & \multicolumn{1}{c|}{3.77} \\ \hline
\multicolumn{1}{l}{} & \multicolumn{1}{l}{} & \multicolumn{1}{l}{} & \multicolumn{1}{l}{} & \multicolumn{1}{l}{} & \multicolumn{1}{l}{} & \multicolumn{1}{l}{} & \multicolumn{1}{l}{} & \multicolumn{1}{l}{} \\ \hline
\multicolumn{1}{|c|}{\multirow{2}{*}{\textbf{Category}}} & \multicolumn{4}{c|}{\textbf{Standard Deviation}} & \multicolumn{4}{c|}{\textbf{Time (s)}} \\ \cline{2-9} 
\multicolumn{1}{|c|}{} & \textbf{U} & \textbf{Lev} & \textbf{Len} & \multicolumn{1}{c|}{\textbf{DEIM}} & \textbf{U} & \textbf{Lev} & \textbf{Len} & \multicolumn{1}{c|}{\textbf{DEIM}} \\ \cline{2-9} 
\multicolumn{1}{|c|}{C2} & 2.62 & 5.64 & 14.19 & \multicolumn{1}{c|}{8.30} & 0.78 & 0.89 & 0.75 & \multicolumn{1}{c|}{0.36} \\
\multicolumn{1}{|c|}{T2} & 0.57 & 6.96 & 5.00 & \multicolumn{1}{c|}{5.58} & 0.70 & 0.78 & 0.68 & \multicolumn{1}{c|}{0.34} \\
\multicolumn{1}{|c|}{O2} & 2.24 & 4.50 & 9.72 & \multicolumn{1}{c|}{4.60} & 0.46 & 0.52 & 0.45 & \multicolumn{1}{c|}{0.22} \\
\multicolumn{1}{|c|}{A2} & 2.25 & 5.92 & 13.04 & \multicolumn{1}{c|}{7.41} & 0.73 & 0.83 & 0.70 & \multicolumn{1}{c|}{0.35} \\
\multicolumn{1}{|c|}{C3} & 1.90 & 2.99 & 15.68 & \multicolumn{1}{c|}{7.17} & 2.47 & 2.77 & 2.35 & \multicolumn{1}{c|}{1.11} \\
\multicolumn{1}{|c|}{T3} & 1.45 & 11.64 & 6.78 & \multicolumn{1}{c|}{8.04} & 1.64 & 1.82 & 1.60 & \multicolumn{1}{c|}{0.78} \\
\multicolumn{1}{|c|}{O3} & 3.16 & 3.74 & 1.82 & \multicolumn{1}{c|}{16.22} & 0.27 & 0.35 & 0.28 & \multicolumn{1}{c|}{0.15} \\
\multicolumn{1}{|c|}{A3} & 1.98 & 6.38 & 14.34 & \multicolumn{1}{c|}{8.64} & 2.18 & 2.44 & 2.08 & \multicolumn{1}{c|}{0.99} \\ \hline
\end{tabular}
\caption{\% Classification Error on Hopkins155 for Algorithm \ref{ALG:RCUR} with $\kappa = 5$ under various sampling methods, (bolded values are best of each category).}
\label{tab:hop-rank-mult-5}
\end{table}

\newpage
\section{Tables of other trials on Yale Extended B}\label{APP:Yale}

\begin{table}[h]
\centering
\begin{tabular}{ccccccccc}
\hline
\multicolumn{1}{|c|}{\multirow{2}{*}{\textbf{Subjects}}} & \multicolumn{4}{c|}{\textbf{Mean}} & \multicolumn{4}{c|}{\textbf{Median}} \\ \cline{2-9} 
\multicolumn{1}{|c|}{} & \textbf{U} & \textbf{Lev} & \textbf{Len} & \multicolumn{1}{c|}{\textbf{DEIM}} & \textbf{U} & \textbf{Lev} & \textbf{Len} & \multicolumn{1}{c|}{\textbf{DEIM}} \\ \cline{2-9} 

\multicolumn{1}{|c|}{2} & 36.98 & \textbf{29.22} & 38.69 & \multicolumn{1}{c|}{47.22} & 36.98 & 29.22 & 38.69 & \multicolumn{1}{c|}{47.22} \\
\multicolumn{1}{|c|}{3} & 50.83 & \textbf{42.55} & 53.88 & \multicolumn{1}{c|}{62.26} & 50.83 & 42.55 & 53.88 & \multicolumn{1}{c|}{62.26} \\
\multicolumn{1}{|c|}{5} & 66.42 & \textbf{58.74} & 67.17 & \multicolumn{1}{c|}{76.12} & 66.42 & 58.74 & 67.17 & \multicolumn{1}{c|}{76.12} \\
\multicolumn{1}{|c|}{8} & 76.72 & \textbf{70.59} & 76.81 & \multicolumn{1}{c|}{84.75} & 76.72 & 70.59 & 76.81 & \multicolumn{1}{c|}{84.75} \\
\multicolumn{1}{|c|}{10} & 80.40 & \textbf{76.07} & 80.23 & \multicolumn{1}{c|}{87.24} & 80.40 & 76.07 & 80.23 & \multicolumn{1}{c|}{87.24} \\
\multicolumn{1}{|c|}{38} & 91.30 & \textbf{91.01} & 91.84 & \multicolumn{1}{c|}{95.07} & - & - & - & \multicolumn{1}{c|}{-} \\ \hline
 &  &  &  &  &  &  &  &  \\ \hline
\multicolumn{1}{|c|}{\multirow{2}{*}{\textbf{Subjects}}} & \multicolumn{4}{c|}{\textbf{Standard Deviation}} & \multicolumn{4}{c|}{\textbf{Time (s)}} \\ \cline{2-9} 
\multicolumn{1}{|c|}{} & \textbf{U} & \textbf{Lev} & \textbf{Len} & \multicolumn{1}{c|}{\textbf{DEIM}} & \textbf{U} & \textbf{Lev} & \textbf{Len} & \multicolumn{1}{c|}{\textbf{DEIM}} \\ \cline{2-9} 
\multicolumn{1}{|c|}{2} & 7.81 & 9.21 & 7.74 & \multicolumn{1}{c|}{3.43} & 0.64 & 3.53 & 0.57 & \multicolumn{1}{c|}{2.98} \\
\multicolumn{1}{|c|}{3} & 6.80 & 8.37 & 5.42 & \multicolumn{1}{c|}{4.29} & 1.29 & 7.39 & 1.32 & \multicolumn{1}{c|}{6.64} \\
\multicolumn{1}{|c|}{5} & 4.49 & 6.61 & 4.29 & \multicolumn{1}{c|}{2.86} & 4.27 & 21.10 & 4.20 & \multicolumn{1}{c|}{20.01} \\
\multicolumn{1}{|c|}{8} & 2.55 & 4.97 & 2.64 & \multicolumn{1}{c|}{1.26} & 15.71 & 61.69 & 16.17 & \multicolumn{1}{c|}{53.29} \\
\multicolumn{1}{|c|}{10} & 2.13 & 3.12 & 2.40 & \multicolumn{1}{c|}{0.82} & 28.92 & 100.92 & 29.54 & \multicolumn{1}{c|}{84.77} \\
\multicolumn{1}{|c|}{38} & - & - & - & \multicolumn{1}{c|}{-} & - & - & - & \multicolumn{1}{c|}{-} \\ \hline
\end{tabular}
\caption{\% Classification Error on Yale Extended B for Algorithm \ref{ALG:RCUR} with $\kappa = 1$ under various sampling methods, (bolded values are best of each category).}
\label{tab:rankmult1}
\end{table}

\begin{table}[h]
\centering
\begin{tabular}{ccccccccc}
\hline
\multicolumn{1}{|c|}{\multirow{2}{*}{\textbf{Subjects}}} & \multicolumn{4}{c|}{\textbf{Mean}} & \multicolumn{4}{c|}{\textbf{Median}} \\ \cline{2-9} 
\multicolumn{1}{|c|}{} & \textbf{U} & \textbf{Lev} & \textbf{Len} & \multicolumn{1}{c|}{\textbf{DEIM}} & \textbf{U} & \textbf{Lev} & \textbf{Len} & \multicolumn{1}{c|}{\textbf{DEIM}} \\ \cline{2-9} 

\multicolumn{1}{|c|}{2} & 9.62 & \textbf{4.46} & 29.05 & \multicolumn{1}{c|}{39.24} & 6.30 & 3.16 & 29.69 & \multicolumn{1}{c|}{45.31} \\
\multicolumn{1}{|c|}{3} & 8.40 & \textbf{6.35} & 35.97 & \multicolumn{1}{c|}{42.80} & 6.77 & 4.83 & 34.13 & \multicolumn{1}{c|}{43.75} \\
\multicolumn{1}{|c|}{5} & \textbf{7.59} & 7.97 & 35.83 & \multicolumn{1}{c|}{46.15} & 6.87 & 6.87 & 33.98 & \multicolumn{1}{c|}{46.28} \\
\multicolumn{1}{|c|}{8} & \textbf{9.81} & 10.23 & 32.83 & \multicolumn{1}{c|}{48.70} & 9.38 & 9.38 & 33.07 & \multicolumn{1}{c|}{49.02} \\
\multicolumn{1}{|c|}{10} & 14.76 & \textbf{10.58} & 33.33 & \multicolumn{1}{c|}{49.36} & 11.98 & 9.59 & 34.75 & \multicolumn{1}{c|}{48.90} \\
\multicolumn{1}{|c|}{38} & \textbf{13.71} & 13.84 & 20.38 & \multicolumn{1}{c|}{52.57} & - & - & - & \multicolumn{1}{c|}{-} \\ \hline
 &  &  &  &  &  &  &  &  \\ \hline
\multicolumn{1}{|c|}{\multirow{2}{*}{\textbf{Subjects}}} & \multicolumn{4}{c|}{\textbf{Standard Deviation}} & \multicolumn{4}{c|}{\textbf{Time (s)}} \\ \cline{2-9} 
\multicolumn{1}{|c|}{} & \textbf{U} & \textbf{Lev} & \textbf{Len} & \multicolumn{1}{c|}{\textbf{DEIM}} & \textbf{U} & \textbf{Lev} & \textbf{Len} & \multicolumn{1}{c|}{\textbf{DEIM}} \\ \cline{2-9} 
\multicolumn{1}{|c|}{2} & 9.36 & 3.89 & 7.79 & \multicolumn{1}{c|}{12.56} & 0.75 & 3.46 & 0.74 & \multicolumn{1}{c|}{2.95} \\
\multicolumn{1}{|c|}{3} & 7.23 & 6.34 & 11.33 & \multicolumn{1}{c|}{12.82} & 1.62 & 7.68 & 1.67 & \multicolumn{1}{c|}{7.03} \\
\multicolumn{1}{|c|}{5} & 5.01 & 6.58 & 9.76 & \multicolumn{1}{c|}{9.51} & 5.07 & 21.94 & 5.13 & \multicolumn{1}{c|}{19.56} \\
\multicolumn{1}{|c|}{8} & 6.36 & 5.61 & 7.13 & \multicolumn{1}{c|}{5.94} & 19.16 & 63.80 & 18.15 & \multicolumn{1}{c|}{55.69} \\
\multicolumn{1}{|c|}{10} & 7.23 & 4.65 & 3.99 & \multicolumn{1}{c|}{3.80} & 34.00 & 107.62 & 34.85 & \multicolumn{1}{c|}{92.31} \\
\multicolumn{1}{|c|}{38} & - & - & - & \multicolumn{1}{c|}{-} & - & - & - & \multicolumn{1}{c|}{-} \\ \hline

\end{tabular}

\caption{\% Classification Error on Yale Extended B for Algorithm \ref{ALG:RCUR} with $\kappa = 2$ under various sampling methods, (bolded values are best of each category).}
\label{tab:rankmult2}
\end{table}

\begin{table}[h]
\centering
\begin{tabular}{ccccccccc}
\hline
\multicolumn{1}{|c|}{\multirow{2}{*}{\textbf{Subjects}}} & \multicolumn{4}{c|}{\textbf{Mean}} & \multicolumn{4}{c|}{\textbf{Median}} \\ \cline{2-9} 
\multicolumn{1}{|c|}{} & \textbf{U} & \textbf{Lev} & \textbf{Len} & \multicolumn{1}{c|}{\textbf{DEIM}} & \textbf{U} & \textbf{Lev} & \textbf{Len} & \multicolumn{1}{c|}{\textbf{DEIM}} \\ \cline{2-9} 
\multicolumn{1}{|c|}{2} & 1.91 & \textbf{1.82} & 5.20 & \multicolumn{1}{c|}{18.87} & 0.79 & 0.82 & 3.13 & \multicolumn{1}{c|}{12.50} \\
\multicolumn{1}{|c|}{3} & \textbf{2.36} & 2.96 & 7.12 & \multicolumn{1}{c|}{20.65} & 1.57 & 2.08 & 4.69 & \multicolumn{1}{c|}{16.15} \\
\multicolumn{1}{|c|}{5} & \textbf{2.55} & 4.37 & 7.92 & \multicolumn{1}{c|}{23.20} & 2.19 & 2.83 & 5.94 & \multicolumn{1}{c|}{20.89} \\
\multicolumn{1}{|c|}{8} & \textbf{2.90} & 5.44 & 9.50 & \multicolumn{1}{c|}{24.45} & 2.76 & 4.30 & 7.81 & \multicolumn{1}{c|}{23.63} \\
\multicolumn{1}{|c|}{10} & \textbf{3.43} & 5.51 & 12.17 & \multicolumn{1}{c|}{25.81} & 3.46 & 5.03 & 9.22 & \multicolumn{1}{c|}{27.83} \\
\multicolumn{1}{|c|}{38} & \textbf{10.40} & 10.56 & 13.01 & \multicolumn{1}{c|}{38.48} & - & - & - & \multicolumn{1}{c|}{-} \\ \hline
 &  &  &  &  &  &  &  &  \\ \hline
\multicolumn{1}{|c|}{\multirow{2}{*}{\textbf{Subjects}}} & \multicolumn{4}{c|}{\textbf{Standard Deviation}} & \multicolumn{4}{c|}{\textbf{Time (s)}} \\ \cline{2-9} 
\multicolumn{1}{|c|}{} & \textbf{U} & \textbf{Lev} & \textbf{Len} & \multicolumn{1}{c|}{\textbf{DEIM}} & \textbf{U} & \textbf{Lev} & \textbf{Len} & \multicolumn{1}{c|}{\textbf{DEIM}} \\ \cline{2-9} 
\multicolumn{1}{|c|}{2} & 2.56 & 2.18 & 6.02 & \multicolumn{1}{c|}{15.10} & 1.13 & 3.89 & 1.18 & \multicolumn{1}{c|}{2.96} \\
\multicolumn{1}{|c|}{3} & 3.42 & 5.26 & 9.75 & \multicolumn{1}{c|}{13.50} & 2.56 & 8.66 & 2.63 & \multicolumn{1}{c|}{6.90} \\
\multicolumn{1}{|c|}{5} & 3.04 & 5.94 & 7.74 & \multicolumn{1}{c|}{10.63} & 7.82 & 24.70 & 7.94 & \multicolumn{1}{c|}{19.88} \\
\multicolumn{1}{|c|}{8} & 1.79 & 4.64 & 5.96 & \multicolumn{1}{c|}{7.42} & 26.05 & 71.32 & 26.62 & \multicolumn{1}{c|}{56.02} \\
\multicolumn{1}{|c|}{10} & 1.08 & 2.82 & 5.88 & \multicolumn{1}{c|}{5.04} & 48.92 & 121.15 & 48.77 & \multicolumn{1}{c|}{94.54} \\
\multicolumn{1}{|c|}{38} & - & - & - & \multicolumn{1}{c|}{-} & - & - & - & \multicolumn{1}{c|}{-} \\ \hline
\end{tabular}
\caption{\% Classification Error on Yale Extended B for Algorithm \ref{ALG:RCUR} with $\kappa = 5$ under various sampling methods, (bolded values are best of each category).}
\label{tab:rankmult5}
\end{table}

\end{document}